\documentclass[10pt,journal,compsoc]{IEEEtran}
\usepackage{graphicx}
\usepackage{amsfonts}

\usepackage{xcolor}
\usepackage{color}
\usepackage{graphicx}
\usepackage{epic}
\usepackage[abs]{overpic}
\usepackage{algorithm}
\usepackage{algorithmic}
\usepackage{amssymb}
\usepackage[cmex10,fleqn]{amsmath}
\usepackage{amsthm}
\usepackage{mathbbold}
\usepackage{mathrsfs}
\usepackage{bbm}
\usepackage{multirow}
\usepackage[T1]{fontenc}
\usepackage{enumitem}
\usepackage{pifont}

\usepackage{arydshln}
\usepackage{subfigure}

\ifCLASSOPTIONcompsoc
  \usepackage[nocompress]{cite}
\else
  \usepackage{cite}
\fi
\ifCLASSINFOpdf
\else
\fi
\begin{document}
%
\title{The Emerging Trends of Multi-Label Learning}
%
%
%
%

\author{Weiwei Liu,
        Haobo Wang,
        Xiaobo Shen,
        and Ivor W. Tsang
\IEEEcompsocitemizethanks{
\IEEEcompsocthanksitem Weiwei Liu is with the School of Computer Science, Wuhan University, Wuhan 430079, China. E-mail: liuweiwei863@gmail.com. \protect\\
\IEEEcompsocthanksitem Haobo Wang is with College of Computer Science and Technology, Zhejiang University. E-mail: wanghaobo@zju.edu.cn. \protect\\
\IEEEcompsocthanksitem Xiaobo Shen is with the School of Computer and Engineering, Nanjing University of Science and Technology, Nanjing 210094, China. E-mail: njust.shenxiaobo@gmail.com. \protect\\
\IEEEcompsocthanksitem Ivor W. Tsang is with the Centre for Artificial Intelligence, FEIT, University of Technology Sydney, NSW, Australia. E-mail: ivor.tsang@uts.edu.au. \protect\\
\IEEEcompsocthanksitem This work is supported by the National Natural Science Foundation of China under Grant No. 61976161, 62176126 and 61906091, the Natural Science Foundation of Jiangsu Province, China (Youth Fund Project) under Grant No. BK20190440, the Fundamental Research Funds for the Central Universities under Grant No. 30921011210, the ARC under Grant No. DP180100106 and DP200101328. \emph{(Corresponding author: Weiwei Liu.)} \protect\\}}

%
%

\markboth{IEEE Transactions on Pattern Analysis and Machine Intelligence}%
{Shell \MakeLowercase{\textit{et al.}}: Bare Demo of IEEEtran.cls for Computer Society Journals}
%



\IEEEtitleabstractindextext{%
\begin{abstract}
Exabytes of data are generated daily by humans, leading to the growing need for new efforts in dealing with the grand challenges for multi-label learning brought by big data. For example, extreme multi-label classification is an active and rapidly growing research area that deals with classification tasks with an extremely large number of classes or labels; utilizing massive data with limited supervision to build a multi-label classification model becomes valuable for practical applications, etc. Besides these, there are tremendous efforts on how to harvest the strong learning capability of deep learning to better capture the label dependencies in multi-label learning, which is the key for deep learning to address real-world classification tasks. However, it is noted that there has been a lack of systemic studies that focus explicitly on analyzing the emerging trends and new challenges of multi-label learning in the era of big data. It is imperative to call for a comprehensive survey to fulfill this mission and delineate future research directions and new applications.
\end{abstract}

\begin{IEEEkeywords}
Extreme Multi-label Learning, Multi-label Learning with Limited Supervision, Deep Learning for Multi-label Learning, Online Multi-label Learning, Statistical Multi-label Learning, New Applications.
\end{IEEEkeywords}}

\maketitle

\IEEEdisplaynontitleabstractindextext

%
\IEEEpeerreviewmaketitle

\IEEEraisesectionheading{\section{Introduction}\label{sec:introduction}}

\IEEEPARstart{M}{ulti-label} classification (MLC), which assigns multiple labels for each instance simultaneously, is of paramount importance in a variety of fields ranging from protein function classification and document classification, to automatic image categorization. For example, an image may have Cloud, Tree and Sky tags; the output for a document may cover a range of topics, such as News, Finance and Sport; a gene can belong to the functions of Protein Synthesis, Metabolism and Transcription.

The traditional multi-label classification methods are not coping well with the increasing needs of today's big and complex data structure. As a result, there is a pressing need for new multi-label learning paradigms and new trends are emerging. This paper aims to provide a comprehensive survey on these emerging trends and the state-of-the-art methods, and discuss the possibility of future valuable research directions.

With the advent of the big data era, extreme multi-label classification (XMLC) becomes a rapidly growing new line of research that focuses on multi-label problems with an extremely large number of labels. Many challenging applications, such as image or video annotation, web page categorization, gene function prediction, language modeling can benefit from being formulated as multi-label classification tasks with millions, or even billions, of labels. The existing MLC techniques can not address the XMLC problem due to the prohibitive computational cost given the large number of labels. One of the most pioneering work in XMLC is SLEEC \cite{DBLP:conf/NeurIPS/BhatiaJKVJ15}, which learns a small ensemble of local distance preserving embeddings. The authors in SLEEC contribute a popular public Extreme Classification Repository \footnote{http://manikvarma.org/downloads/XC/XMLRepository.html}, which promote the development of XMLC. The state-of-the-art XMLC techniques are mostly based on one-vs-all classifiers \cite{DBLP:conf/wsdm/BabbarS17,DBLP:conf/kdd/YenHDRDX17,DBLP:conf/www/PrabhuKHAV18,DBLP:conf/wsdm/JainBCV19}, trees \cite{DBLP:conf/kdd/PrabhuV14,DBLP:conf/kdd/JainPV16,DBLP:conf/icml/JasinskaDBPKH16,DBLP:conf/wsdm/PrabhuKGDHAV18,DBLP:conf/NeurIPS/WydmuchJKBD18} and embeddings \cite{DBLP:conf/icml/Yu0KD14,DBLP:conf/NeurIPS/BhatiaJKVJ15,DBLP:conf/kdd/Tagami17,DBLP:journals/pami/LiuXTZ19,DBLP:conf/ijcai/GongYB21}. Unfortunately, the theoretical results in XMLC under the very high dimensional settings remain relatively under-explored. Moreover, the labels are extremely sparse, which leads to the problem of the long-tail distribution. How to precisely predict all the positive labels to testing examples pose a serious challenge in XMLC.

As the data volume grows quickly these days, it is usually expensive and time-consuming to acquire full supervision. In MLC tasks, the high dimensional output space makes it even harder. To mitigate this problem, a wealth of works have proposed various settings of MLC with limited supervision. For example, multi-label learning with missing labels (MLML) \cite{DBLP:conf/aaai/SunZZ10} assumes that only a subset of labels is obtained; semi-supervised MLC (SS-MLC) \cite{DBLP:conf/sdm/ChenSWZ08} admits a few fully labeled data and a large amount of unlabeled data; partial multi-label learning (PML) \cite{DBLP:conf/aaai/XieH18} studies an ambiguous setting that a superset of labels is given. Many effective models are also proposed based on graph \cite{DBLP:conf/aaai/SunZZ10,DBLP:conf/ijcai/Wang0ZZHC19,huynh2020interactive}, embedding \cite{DBLP:conf/icml/Yu0KD14,DBLP:conf/kdd/XuT016,DBLP:conf/aaai/SunFWLJ19}, probability models \cite{DBLP:conf/icml/JainMR17,DBLP:conf/eccv/ChuYW18} and so on. More interesting improperly-supervised MLC settings are also considered recently, such as MLC with noisy labels \cite{DBLP:conf/cvpr/HuHSC19}, multi-label zero-shot learning \cite{ji2020deep} and multi-label active learning \cite{DBLP:conf/icml/Shi019}. These settings make MLC practical in real-world applications by saving supervision costs, and thus, deserve more attention.

Deep learning has shown excellent potential since 2012 when AlexNet presents surprising performance on the single-label image classification of ILSVRC2012 challenge.
As most natural images usually contain multiple objects, it is more practical that each image is associated with multiple tags or labels.
Thus developing deep learning techniques that can address MLC problem is more practically demanding in real-world image classification tasks.
Some large-scale multi-label image databases, e.g., Open Images \cite{DBLP:journals/corr/abs-1811-00982}, newly released Tencent ML-Images \cite{DBLP:journals/access/WuCFZHLZ19} promote deep learning for MLC problem.
In this area, BP-MLL  \cite{Zhang:2006:MNN:1159162.1159294} is the first method to utilize neural network (NN) architecture for MLC problem.
Canonical Correlated AutoEncoder (C2AE)  \cite{DBLP:conf/aaai/YehWKW17} is the first Deep NN (DNN) based embedding method for MLC problem.
In addition, some deep learning methods are also developed for the Challenging MLC problems, such as Extreme MLC \cite{DBLP:conf/sigir/LiuCWY17,DBLP:conf/sigir/LiuCWY17,DBLP:conf/NeurIPS/YouZWDMZ19,DBLP:conf/naacl/WangCSQLZ19}, partial and weakly-supervised MLC \cite{DBLP:conf/cvpr/DurandMM19,huynh2020interactive,DBLP:conf/eccv/ChuYW18,DBLP:conf/eccv/ChuYW18}, MLC with unseen labels \cite{Wang2020,DBLP:conf/cvpr/LeeFYW18,DBLP:conf/cvpr/LeeFYW18}.
Recently advanced deep learning architectures \cite{DBLP:conf/icml/CisseAB16,DBLP:conf/NeurIPS/NamMKF17,DBLP:conf/icml/NamKMPSF19,DBLP:journals/corr/abs-1911-06557} for MLC problems are studied.
How to harvest the strong learning capability of deep learning to better capture the label dependencies is key for deep learning to address MLC problems.

The Web continues to generate quintillion bytes of streaming data daily, leading to the key challenges for MLC tasks. Firstly, the existing off-line MLC algorithms are impractical for streaming data sets, since they require to store all data sets in memory; secondly, it is non-trivial to adapt off-line multi-label methods to the sequential data. Therefore, several approaches for online multi-label classification have recently been proposed, including \cite{DBLP:conf/icassp/ParkC13,DBLP:journals/evs/VenkatesanEDPW17,DBLP:conf/aaai/GongYB20}. However, both the experimental and theoretical results obtained so far are still not satisfactory and very limited. There is a real pressing need for credible research into online multi-label learning.
\begin{figure}[t]
  \centering
  \includegraphics[width=0.99\linewidth]{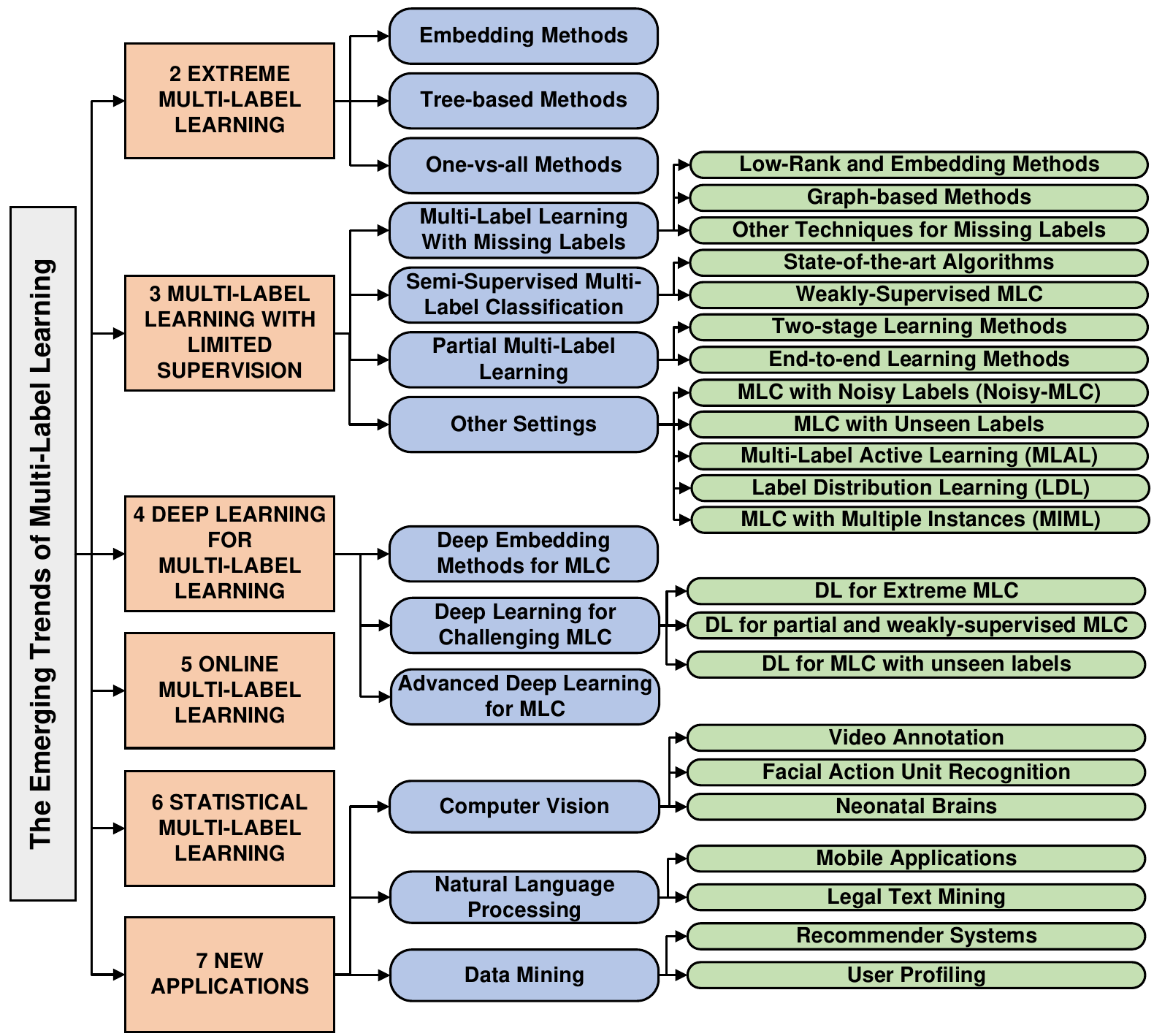}\\
  \caption{The structure of this paper.}\label{Fig:structure}
\end{figure}
Many references \cite{journals/pr/ZhangZ07,JMLR:v18:16-212} have shown that methods of multi-label learning which explicitly capture label dependency will usually achieve better prediction performance.
Therefore, in the past few years, modeling the label dependency is one of the major challenges in multi-label classification problems. A plethora of methods have been motivated to model the dependence. For example, the classifier chain (CC) model \cite{Read:2009:CCM:1617459.1617477} captures label dependency by using binary label predictions as extra input attributes for the following classifiers in a chain. CCA \cite{journals/jmlr/ZhangS11a} uses canonical correlation analysis for learning label dependency. CPLST \cite{DBLP:conf/NeurIPS/ChenL12} uses principal component analysis to capture both the label and the feature dependencies. Unfortunately, the statistical properties and asymptotic analysis of all these methods are still not well explored. One of the emerging trends is to develop statistical theory for understanding multi-label dependency modelings.

During the past decade, multi-label classification has been successfully applied in computer vision, natural language processing and data mining. This paper will briefly review these emerging applications, which may inspire the community to explore more interesting applications.
The structure of this paper is shown in Figure \ref{Fig:structure}. Some evaluation metrics and important notations used in this paper can be found in the Supplementary Materials.

\section{Extreme Multi-label Learning}\label{xml}

Extreme multi-label classification (XMLC) aims to learn a classifier that is able to automatically annotate a data point with the most relevant subset of labels from an extremely large number of labels, which has opened up a new research frontier in data mining and machine learning. For example, there are millions of people who upload their selfies on the Facebook every day, based on these selfies, one might wish to build a classifier that recognizes who appear in the figure.
Many XMLC applications have been found in various domains ranging from language modeling, document classification and face recognition to gene function prediction. The main challenging issue of XMLC is that XMLC learns with hundreds of thousands, or even millions, of labels, features and training points. To address this issue, the state-of-the-art XMLC techniques are mostly based on embeddings, trees and one-vs-all classifiers. We will review these advanced techniques in this section. Note that there are also some new deep learning-based XMLC models, but we leave the discussion until \S \ref{MLCwithDL}.


\subsection{Embedding Methods}\label{embedding-models}

To deal with many labels, \cite{conf/NeurIPS/HsuKLZ09} assume that label vectors have a little support. In other words, each label vector can be projected into a lower dimensional compressed label space, which can be deemed as encoding. A regression is then learned for each compressed label. Lastly, the compressed sensing technique is used to decode the labels from the regression outputs of each testing instance. Many embedding based works have recently been developed in this learning paradigm. These works mainly differ in compression and decompression methods such as canonical correlation analysis (CCA) \cite{journals/jmlr/ZhangS11a} and bloom filters \cite{DBLP:conf/NeurIPS/CisseUAG13}. Amongst them, SLEEC \cite{DBLP:conf/NeurIPS/BhatiaJKVJ15} is one of the seminal embedding methods in XMLC due to its simplicity and promising experimental results \cite{DBLP:conf/NeurIPS/BhatiaJKVJ15}.

SLEEC learns low dimensional embeddings which non-linearly capture label correlations by preserving the pairwise distances
between only the closest (rather than all) label vectors. Regressors are then trained in the embedding space. SLEEC uses a $k$-nearest neighbor ($k$NN) classifier in the embedding space for prediction.

Assume $x_i \in \mathbb{R}^{d \times 1} $ is a real vector representing an input or instance (feature), $y_i \in \{0,1\}^{L \times 1}$ is the corresponding output or label vector $(i \in \{1,\ldots,n\})$. $n$ denotes the number of training data. The input matrix is $X=[x_1,\ldots,x_n] \in \mathbb{R}^{d\times n} $ and the output matrix is $Y=[y_1,\ldots,y_n] \in \{0,1\}^{L \times n} $.
SLEEC maps the label vector $y_i$ to $\varpi$-dimensional vector $z_i \in \mathbb{R}^{\varpi \times 1}$ ($\varpi < L$ is a small constant) and learns a set of regressors $V \in \mathbb{R}^{\varpi \times d} $ s.t. $z_i\approx V x_i, \forall i \in \{1,\ldots,n\}$. During the prediction, for a testing instance $x$, SLEEC first computes its embedding $Vx$ and then perform $k$NN over the set $[Vx_1,\ldots,Vx_n]$.
We denote the transpose of the vector/matrix by the superscript $^T$ and the logarithms to base 2 by $\log$. Let $||\cdot ||_F$ and $||\cdot||_1$ represent the Frobenius norm and $\ell_1$ norm of a matrix.

SLEEC aims to learn a embedding matrix $Z=[z_1,\ldots,z_n] \in \mathbb{R}^{\varpi \times n} $ through the following formula:
\begin{equation}\label{1}
\min_{Z\in \mathbb{R}^{\varpi \times n}} ||P_{\Omega}(Y^TY) - P_{\Omega}(Z^TZ)||^2_F
\end{equation}
where the index set $\Omega$ denotes the set of neighbors: $(i,j)\in \Omega$ iff $j \in \mathcal {N}_i$. $\mathcal {N}_i$ denotes a set of nearest neighbors of $i$. $P_{\Omega}(\cdot)$ is defined as:
\begin{equation}\label{2}
\begin{split}
\big(P_{\Omega}(Y^TY)\big)_{(i,j)}=
\begin{cases}
   y_i^Ty_j, &\mbox{if $(i,j)\in \Omega$}\\
   0, &\mbox{otherwise.}
\end{cases}
\end{split}
\end{equation}
Based on embedding matrix $Z$, SLEEC minimizes the following objective with $\ell_1$ and $\ell_2$ regularization to find regressors $V$, which is able to reduce the prediction time and the model size, and avoid overfitting.
\begin{equation}\label{3}
\begin{split}
\min_{V \in \mathbb{R}^{\varpi \times d} } ||Z - VX||^2_F + \mu ||V||^2_F+ \lambda||VX||_1
\end{split}
\end{equation}
where $\mu > 0$ and $\lambda > 0$ are the regularization parameters.

To scale to large-scale data sets, SLEEC clusters the training set into smaller local region just based on features and does not consider label information. Therefore, the instances that have similar labels are not guaranteed to be split into the same region. This partitioning may affect the quality of embeddings learned in SLEEC.

Many methods have been developed to address this issue. For example, AnnexML \cite{DBLP:conf/kdd/Tagami17} shows a novel graph embedding method based on the k-nearest neighbor graph (KNNG). AnnexML aims to construct the KNNG of label vectors in the embedding space to improve both the prediction accuracy and speed of the k-nearest neighbor classifier. DEFRAG \cite{DBLP:conf/ijcai/JalanK19} represents each feature $j\in [d]$ as an $L$-dimensional vector $q^j=\sum_{i=1}^n x_i^j y_i$, which is a weighted aggregate of the label vectors of data points where the feature $j$ is non-zero. After creating these representative vectors, DEFRAG performs hierarchical clustering on them to obtain feature clusters, and then performs agglomeration by summing up the coordinates of the feature vectors within a cluster. \cite{DBLP:conf/ijcai/JalanK19} shows that DEFRAG offers faster and better performance.

Word embeddings have been successfully used for learning non-linear representations of text data for natural language processing (NLP) tasks, such as understanding word and document semantics and classifying documents. Recently, \cite{DBLP:conf/aaai/0001WNK0R19} first proposes to use word embedding techniques to learn the label embedding of instances. \cite{DBLP:conf/aaai/0001WNK0R19} treats each instance as a ``word'', and define the ``context'' as k-nearest neighbors of a given instance in the space formed by the training label vectors $y_i$. Based on Skip Gram Negative Sampling (SGNS) technique, \cite{DBLP:conf/aaai/0001WNK0R19} learns embeddings $z_1,\ldots,z_n$ through the following formula:
\begin{equation}\label{4}
\max_{z_1,\ldots,z_n} \sum_{i=1}^n  \bigg( \sum_{j \in \mathcal {N}_i} \log(\sigma (\langle z_i,z_j \rangle)) + C \sum_{j'} \log(\sigma (-\langle z_i,z_{j'} \rangle)) \bigg)
\end{equation}
where $j'\in \{1,\ldots,n\}$, $\sigma (\cdot)$ is a sigmoid function, $\langle \cdot,\cdot \rangle$ denotes the inner product and $C$ is a constant. After learning label embeddings $z_1,\ldots,z_n$, \cite{DBLP:conf/aaai/0001WNK0R19} follows the learning algorithm of SLEEC to learn $V$ and make the prediction. \cite{DBLP:conf/aaai/0001WNK0R19} shows competitive prediction accuracies compared to state-of-the-art embedding methods, and provides the new insight for XMLC from the popular word2vec in NLP, which may open a new line of research.

The embedding matrix $Z=[z_1,\ldots,z_n] \in \mathbb{R}^{\varpi \times n}$ of existing embedding methods is in real space. Hence we need to use regressors for training and may involve solving expensive optimization problems. To break this limitation, many references leverage coding technique for efficiently training the model. For example, based on Bloom filters \cite{DBLP:journals/cacm/Bloom70}, a well-known space-efficient randomized data structure designed for approximate membership testing, \cite{DBLP:conf/NeurIPS/CisseUAG13} designs a simple scheme to select the $k$ representative bits for labels for training and proposes a robust decoding algorithm for prediction. However, Bloom filters may yield many false positives. 

To address this issue, \cite{DBLP:conf/icml/UbaruM17} transforms MLC to a popular group testing problem. In the group testing problem, one wish to identify a small number $k$ of defective elements in a population of large size $L$. The idea is to test the items in groups with the premise that most tests will return negative results. Only few $\varpi < L$ tests are needed to detect the $k$ defective elements. 
\cite{DBLP:conf/icml/UbaruM17} trains $\varpi$ binary classifiers on $z_i$ and learn to test whether the data belongs to a group (of labels) or not, and then uses a simple inexpensive decoding scheme to recover the label vector from the predictions of the classifiers. Recently, \cite{JMLR:v18:16-466} develops a novel sparse coding tree framework for XMLC based on Huffman coding and Shannon-Fano coding. \cite{DBLP:conf/NeurIPS/CisseUAG13,DBLP:conf/icml/UbaruM17,JMLR:v18:16-466} introduce the coding theory into MLC which is very novel and worth further research and exploration in this direction.

\textbf{Remark.} Embedding methods are the most popular strategies for addressing XMLC. SLEEC is a seminal work among them and recommended for the beginners to try. The major limitation of existing embedding methods is that the correlations between the input and output are ignored, such that their learned embeddings are not well aligned, which leads to degradation in prediction performance. How to build an embedding space that can preserve the relations between the input and output is an important research topic in the future. For example, \cite{DBLP:conf/aaai/YehWKW17,DBLP:journals/tnn/ShenLTSO18,DBLP:conf/acml/WangYYY18} explore the correlations between the input and output. They propose to jointly learn a semantic common subspace and view-specific mappings within one framework. The semantic similarity structure among the embeddings is further preserved, ensuring that close embeddings share similar labels. Another limitation of existing embedding methods is that both the training and testing time complexity are too high (See Table \ref{Timecost}). Some techniques, such as random projection, hashing and parallelization, may be able to accelerate the training and testing process. Tree-based methods are able to obtain fast testing speed, which is discussed in the following paragraph.

\subsection{Tree-based Methods}
For tree-based methods, the original large-scale problem is divided into a sequence of small-scale subproblems by hierarchically partitioning the instance set or the label set. The root node is initialized to contain the entire set. A partitioning formulation is then optimized to partition a set in a node into a fixed number $k$ of subsets which are linked to $k$ child nodes. Nodes are recursively decomposed until a stopping condition is checked on the subsets. Each node involves two optimization problems: optimizing the partition criterion, and defining a condition or building a classifier on the feature space to decide which child node an instance belongs to. In the prediction phase, an instance is passed down the tree until it reaches a leaf (instance tree) or several leaves (label tree). For a label tree, the reached leaves contain the predicted labels. For an instance tree, the prediction is made by a classifier trained on the instances in the leaf node. Thus, the main advantage of tree-based methods is that the prediction costs are sub-linear or even logarithmic if the tree is balanced.

FastXML \cite{DBLP:conf/kdd/PrabhuV14} presents to learn the hierarchy by optimizing the ranking loss function, normalized Discounted Cumulative Gain (nDCG). nDCG brings two main benefits to XMLC. Firstly, nDCG is a measurement which is sensitive to ranking and relevance and therefore ensures that the relevant positive labels are predicted with ranks that are as high as possible. This cannot be guaranteed by rank insensitive measures such as the Gini index or the clustering error. Second, by being rank sensitive, nDCG can be optimized across all $L$ labels at the current node thereby ensuring that the local optimization is not myopic. The experiments show that nDCG is more suitable for extreme multi-label learning.

Based on FastXML, PfastreXML \cite{DBLP:conf/kdd/JainPV16} studies how to improve the prediction accuracy of tail labels. The labels in XMLC follow a power law distribution. Infrequently occurring labels usually convey more information, but have little training data and are harder to predict than frequently occurring ones. PfastXML improves FastXML by replacing the nDCG loss with its unbiased propensity scored variant, and assigns higher rewards for predicting accurate tail labels. Moreover, PfastreXML re-ranks PfastXML's predictions using tail label classifiers. \cite{DBLP:conf/kdd/JainPV16} shows that PfastreXML achieves promising performance in predicting tail labels and successfully applies to tagging, recommendation and ranking problems. SwiftXML \cite{DBLP:conf/wsdm/PrabhuKGDHAV18} maintains all the scaling properties of PfastreXML, but improves the prediction accuracy of PfastreXML by considering more information about revealed item preferences and item features. SwiftXML proposes a novel node partitioning function by optimizing two separating hyperplanes in the user and item feature spaces respectively. Experiments on tagging on Wikipedia and item-to-item recommendation on Amazon reveal that SwiftXML is more accurate than leading extreme classifiers by 14\%.

\begin{table}[t]
\caption{The training and testing time complexity of XMLC methods ($\text{nnz}(X)$ denotes the number of non-zeros in $X$, $C$ is a constant, $O(\zeta)$ denotes the computational complexity of $\varpi$-bit Hamming distance calculation. $T$ is the number of trees. $h$ is the number of levels in the tree. $c$ is the number of top-scoring items being reranked by the base-classifiers. $k\ll L$ is a small constant).}
\label{Timecost}
\begin{center}
\tiny
\begin{tabular}{llll}
\multicolumn{1}{c}{\bf Methods} & \multicolumn{1}{c}{\bf Training Time} & \multicolumn{1}{c}{\bf Testing Time}
\\ \hline \\
Embedding: SLEEC \cite{DBLP:conf/NeurIPS/BhatiaJKVJ15} &$O(n\varpi^2+n\varpi C)$ & $O(nd+kL)$\\
Embedding: DEFRAG \cite{DBLP:conf/ijcai/JalanK19} &$O(\text{nnz}(X)\log d)$ & $O(nd+kL)$\\
Embedding: CoH \cite{DBLP:journals/tnn/ShenLTSO18} &$O(n(d^2+L^2))$ & $O(n\zeta+kL)$\\ \cline{1-4}
Tree: FastXML \cite{DBLP:conf/kdd/PrabhuV14} &$O(nT\log L+ \text{nnz}(X)nT )$ & $O(Td\log L)$\\
Tree: SwiftXML \cite{DBLP:conf/wsdm/PrabhuKGDHAV18}&$O(\text{nnz}(X) Tn\log n  )$ & $O((T\log n \!+\!c)\text{nnz}(X))$\\
Tree: GBDT-SPARSE \cite{DBLP:conf/icml/SiZKMDH17}&$O(\text{nnz}(X) dTh\log k  )$ & $O(Tk\log k)$\\ \cline{1-4}
OVA: PD-Sparse \cite{DBLP:conf/icml/YenHRZD16}&$O(ndC )$ & $O(dL)$\\
OVA: LF \cite{DBLP:conf/aistats/Niculescu-Mizil17} &$O(nd\!+ \!L\log L\!+\!\! n\log n\!+\!nL )$ & $O(d+\log (2L))$\\
OVA: Parabel \cite{DBLP:conf/www/PrabhuKHAV18} &$O((nd\log L)/L)$ & $O(\text{nnz}(X) Tk\log L )$\\
OVA: Slice \cite{DBLP:conf/wsdm/JainBCV19} &$O(nd\log L)$ & $O(d\log L )$\\
\hline
\end{tabular}
\end{center}
\end{table}
FastXML, PfastreXML and SwiftXML have studied the ranking-based measures such as nDCG and its variants. Recently, \cite{DBLP:conf/icml/JasinskaDBPKH16} focuses on F-measure, which is a commonly used performance measure in multi-label classification as well as other fields, such as information retrieval and natural language processing.
\cite{DBLP:conf/icml/JasinskaDBPKH16} proposes a novel sparse probability estimates (SPEs) to reduce the complexity of threshold tuning in XMLC. Then, they develop three algorithms for maximizing the F-measure in the Empirical Utility Maximization (EUM) framework by using SPEs. Moreover, Probabilistic label trees (PLTs) and FastXML are discussed for computing SPEs. Recently, the theory in \cite{DBLP:conf/NeurIPS/WydmuchJKBD18} shows that the pick-one-label is inconsistent with respect to the Precision@$k$, and PLTs model can get zero regret (i.e., it is consistent) in terms of marginal probability estimation and Precision@$k$ in the multi-label setting. Inspired by \cite{DBLP:conf/NeurIPS/WydmuchJKBD18}, \cite{DBLP:conf/NeurIPS/X19} further studies the consistency of other reduction strategies based on a different Recall@$k$ metric.

\textbf{Remark.} Tree-based methods are the efficient strategy for addressing XMLC with the logarithmic dependence to the number of labels (See Table \ref{Timecost}). FastXML is a popular method and recommended for the practitioners. However, one of the major problems for tree-based methods, such as FastXML, PfastreXML and SwiftXML, is that they involve complex non-convex optimization problem at each node. How to obtain cheap and scalable tree structure is an important research topic in the future. For example, GBDT-SPARSE \cite{DBLP:conf/icml/SiZKMDH17} studies the gradient boosted decision trees (GBDT) for XMLC. In each node, the feature is firstly projected into a low-dimensional space and then a simple inexact search strategy is used to find a good split. They significantly reduce the training and prediction time and model size of GBDT to make it suitable for XMLC. CRAFTML \cite{DBLP:conf/icml/SibliniMK18} tries to use fast partitioning strategies and exploit random forest algorithm. CRAFTML first randomly projects the feature and label into lower dimensional spaces. A $k$-means algorithm is then used in the projected labels to partition the instances into $k$ temporary subsets. Moreover, GBDT-SPARSE and CRAFTML also open the way to parallelization, which are able to motivate further research.

\subsection{One-vs-all Methods}

One-vs-all (OVA) methods are one of the most popular strategies for multi-label classification which independently trains a binary classifier for each label. However, this technique suffers two major limitations for XMLC: 1) Training one-vs-all classifiers for XMLC problems using off-the-shelf solvers such as Liblinear can be infeasible for computation and memory. 2) The model size for XMLC data set can be extremely large, which leads to slow prediction. Recently, many works have been developed to address the above issues of the one-vs-all methods in XMLC.

By exploiting the sparsity of the data, some sub-linear algorithms are proposed to adapt one-vs-all methods in the extreme classification setting. For example, PD-Sparse \cite{DBLP:conf/icml/YenHRZD16} proposes to minimize the separation ranking loss and $\ell_1$ penalty in an Empirical Risk Minimization (ERM) framework for XMLC. The separation ranking loss penalizes the prediction on an instance by the highest response from the set of negative labels minus the lowest response from the set of positive labels. PD-Sparse obtains an extremely sparse solution both in primal and in dual with the sub-linear time cost, while yields higher accuracy than SLEEC, FastXML and some other XMLC methods. By introducing separable loss functions, PPDSparse \cite{DBLP:conf/kdd/YenHDRDX17} parallelizes PD-Sparse with sub-linear algorithms to scale out the training. PPDSparse can also reduce the memory cost of PDSparse by orders of magnitude due to the separation of training for each label. DiSMEC \cite{DBLP:conf/wsdm/BabbarS17} also presents a sparse model with a parameter thresholding strategy, and employs a double layer of parallelization to scale one-vs-all methods for problems involving hundreds of thousand labels. ProXML \cite{DBLP:journals/ml/BabbarS19} proposes to use $\ell_1$-regularized Hamming loss to address the tail label issues, and reveals that minimizing one-vs-all method based on Hamming loss works well for tail-label prediction in XMLC based on the graph theory.

PD-Sparse, PPDSparse, DiSMEC and ProXML have obtained high prediction accuracies and low model sizes. However, they still train a separate linear classifier for each label and linear scan every single label to decide whether it is relevant or not. Thus the training and testing cost of these methods grow linearly with the number of labels. Some advanced methods are presented to address this issue. For example, to reduce the linear prediction cost of one-vs-all methods, \cite{DBLP:conf/aistats/Niculescu-Mizil17} proposes to predict on a small set of labels, which is generated by projecting a test instance on a filtering line, and retaining only the labels that have training instances in the vicinity of this projection. The candidate label set should keep most of the true labels of the testing instances, and be as small as possible. They train the label filters by optimizing these two principles as a mixed integer problem. The label filters can reduce the testing time of existing XMLC classifiers by orders of magnitude, while yields comparable prediction accuracy. \cite{DBLP:conf/aistats/Niculescu-Mizil17} shows an interesting technique to find a small number of potentially relevant labels, instead of going through a very long list of labels. How to use label filters to speed up the training time is left as an open problem.

Parabel \cite{DBLP:conf/www/PrabhuKHAV18} reduces training time of one-vs-all methods from $O(ndL)$ to $O((nd\log L)/L)$ by learning balanced binary label trees based on an efficient and informative label representation. They also present a probabilistic hierarchical multi-label model for generalizing hierarchical softmax to the multi-label setting. The logarithmic prediction algorithm is also proposed for dealing with XMLC. Experiments show that Parabel could be orders of magnitude faster at training and prediction compared to the state-of-the-art one-vs-all extreme classifiers. However, Parabel is not accurate in low-dimension data set, because Parabel can not guarantee that similar labels are divided into the same group, and the error will be propagated in the deep trees. To reduce the error propagation, Bonsai \cite{DBLP:journals/corr/abs-1904-08249} shows a shallow $k$-ary label tree structure with generalized label representation. A novel negative sampling technique is also presented in Slice \cite{DBLP:conf/wsdm/JainBCV19} to improve the prediction accuracy for low-dimensional dense feature representations. Slice is able to cut down the training time cost of one-vs-all methods from linear to $O(nd\log L)$ by training classifier on only $O(n/L\log L)$ of the most confusing negative examples rather than on all $n$ training set. Slice employs generative model to estimate $O(n/L\log L)$ negative examples for each label based on approximate nearest neighbor search (ANNS) in time $O((n+L)d\log L)$, and conduct the prediction on $O(\log L)$ of the most probable labels for each testing data. Slice is up to 15\% more accurate than Parabel, and able to scale to 100 million labels and 240 million training points. The experiments in \cite{DBLP:conf/wsdm/JainBCV19} show that negative sampling is a powerful tool in XMLC, and the performance gain of some advanced negative sampling technique may be explored for future research.

\textbf{Remark.} One-vs-all methods are the simple strategies for dealing with XMLC, and PD-Sparse is the first choice for the beginners to try. As mentioned before, one-vs-all methods independently train a binary classifier for each label, so computation and memory cost pose an intractable issue for XMLC, and one-vs-all methods do not consider the correlations between labels. Although the reviewed methods in this subsection are able to ease the computation issue, how to use the correlations between labels to boost the performance of one-vs-all methods could pose a serious problem in the future. One possible way is to design some one-vs-all learning models which consider various label correlations.


\section{Multi-label Learning With Limited Supervision}\label{MLCwithlimitedsup}
Collecting fully-supervised data is usually hard and expensive and thus a critical bottleneck in real-world classification tasks. In MLC problems, there exist many ground-truth labels and the output space can be very large, which further aggravates the difficulty of precise annotation. To mitigate this problem, plenty of works have studied different settings of MLC with limited supervision. How to model label dependencies and handle incomplete supervision pose two major challenges in these tasks. In this section, we concentrate on several advanced topics. Amongst them, multi-label learning with missing labels (MLML) assumes only a subset of labels are given; semi-supervised MLC (SS-MLC) assumes a large set of unlabeled data as well labeled data are given; partial multi-label learning (PML) allows the annotators to provide a superset of labels as the candidates. We illustrate the connections between these different supervision types in Figure \ref{wsmlc-settings}. Note that in these settings, though trained with imperfect supervised signals, the classifier is still evaluated on a perfectly supervised testing data set to quantify the predictive performance.

\subsection{Multi-Label Learning With Missing Labels}
\begin{figure*}
	\centering
	\subfigure[Supervised MLC]{
	   \includegraphics[width=0.15\linewidth]{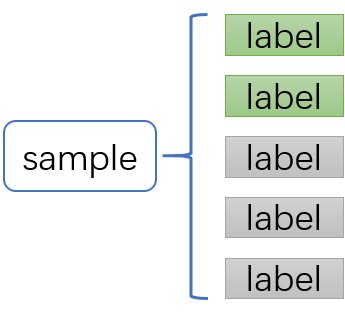}
    	}\hspace{8mm}
	\subfigure[Explicit MLML]{
	   \includegraphics[width=0.15\linewidth]{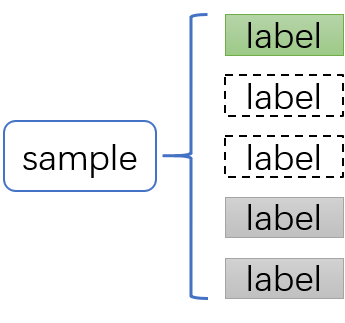}
    	}\hspace{8mm}
    \subfigure[Implicit MLML]{
	   \includegraphics[width=0.15\linewidth]{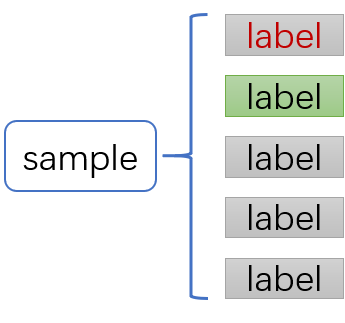}
    	}\hspace{8mm}
	\subfigure[PML]{
	   \includegraphics[width=0.15\linewidth]{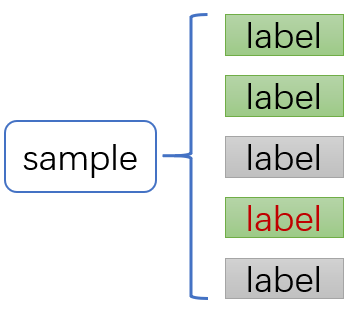}
    	}\\
    \includegraphics[width=0.68\linewidth]{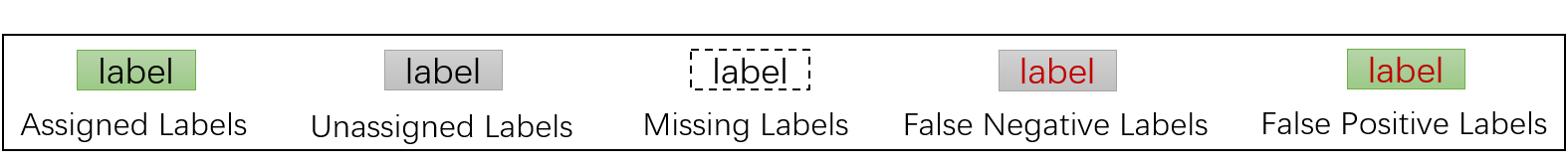}
	\caption{Illustration of some MLC settings with different types of supervision. (a) instances with full supervision; (b) instances with explicitly missing labels; (c) instances with implicitly missing labels; (d) instances with a set of candidate relevant labels. Here (b) and (c) are two different settings of MLML problems. Semi-supervised MLC is a special case of (b) where some instances miss all the labels.}
	\label{wsmlc-settings}
\end{figure*}

In real-world scenarios, it is intractable for the annotators to figure out all the ground-truth labels, due to the complicated structure or the high volume of the output space. Instead, a subset of labels can be obtained, which is called multi-label learning with missing labels (MLML). There are two main settings in MLML. The first setting \cite{DBLP:conf/aaai/SunZZ10} only obtains a subset of relevant labels. It views the MLML problem as a positive-unlabeled learning task such that the remaining labels are all regarded as negative labels. The other setting \cite{DBLP:conf/icpr/WuLWHJ14} explicitly indicates which labels are missing. Formally, given a feature vector $x_i$, we denote the label vector of these two settings by $\hat{y}_i\in\{-1, +1\}^{L\times1}$ ($-1$ can be missing or negative labels) and $\tilde{y}_i\in\{-1, 0, +1\}^{L\times1}$ ($0$ represents missing labels) respectively. We distinguish these two settings in Figure \ref{wsmlc-settings}. Moreover, two different learning targets may be considered. One is transductive that only learns to complete the missing entries. The other is inductive where a classifier is trained for unseen data. For simplicity, we do not explicitly distinguish these differences.

Next, we will review state-of-the-art MLML methods which are mainly based on low-rank and graph assumptions.

\subsubsection{Low-Rank and Embedding Methods}
As discussed in \S \ref{embedding-models}, the existence of label correlations usually implies the output space is low-rank. Interestingly, this assumption has been widely used to complement the missing entries of a matrix in matrix completion tasks \cite{DBLP:conf/NeurIPS/XuJZ13}. Since it benefits the two key targets in MLML, i.e. label correlation extraction and missing label completion, many low-rank assumption-based MLML methods have been developed. 

In \cite{DBLP:conf/NeurIPS/XuJZ13}, the MLML problem is regarded as a low-rank matrix completion problem with the existence of side information, i.e. the features. To accelerate the learning task, the label matrix is decomposed to be $Y=AWB$ where $A$ and $B$ are side information matrices. $W$ is assumed to be low-rank. In fact, in MLML problems, $A$ is exactly the feature matrix $X$ and $B$ is the identity matrix since there is no side information for the labels. Therefore, $W$ can be viewed as a linear classifier that enables the predicted labels $Y=XW$ to be low-rank. Then, LEML \cite{DBLP:conf/icml/Yu0KD14} generalizes this paradigm to a flexible empirical risk minimization framework. The formula is as follows,
\begin{equation}\label{LEML}
\begin{split}
W=\arg\min_W \mathcal{L}(\hat{Y}, XW) + \lambda r(W), \quad \textrm{s.t. } \textrm{rank}(W)\le k
\end{split}
\end{equation}
where $\lambda$ and $k$ are constants, $r(W)$ is the regularizer. $\mathcal{L}$ can be any empirical risk that is evaluated on observed entries. To solve this problem, \cite{DBLP:conf/icml/Yu0KD14} decomposes the classifier to two rank-$k$ ($k\ll L$) matrices $V$ and $U$ such that $W=VU$. Then, an alternative optimization method is used to efficiently handle large-scale problems. Nevertheless, the presence of tail labels may break the low-rank property. Hence, \cite{DBLP:conf/kdd/XuT016} treats the tail labels as outliers and decompose the label matrix to, $\hat{Y}\approx Y_1 + Y_2$. Here $Y_1$ is low-rank and $Y_2$ is sparse. These two components can be obtained by solving the following objective,
\begin{equation}\label{REML}
\begin{split}
\min_{U,V,H} ||\hat{Y}-Y_1-&Y_2||_F^2 + \lambda_1||H||_F^2\\
&+\lambda_2(||U||_F^2+||V||_F^2)+\lambda_3||XH||_1 \\
\quad \textrm{s.t.}\quad Y_1=& XUV, \quad  Y_2=XH
\end{split}
\end{equation}
These two learning frameworks are followed by many works. For example, \cite{DBLP:conf/kdd/HanSSZ18} studies the problem that both features and labels are incomplete. The proposed solution, ColEmbed, requires the classifier as well as the recovered feature matrix to be low rank. Moreover, the kernel trick is used to enable the non-linearity of the classifier. Some recent works \cite{DBLP:conf/eccv/YangZC16,DBLP:conf/aaai/YuHDL17,DBLP:journals/ijcv/WuJLGL18} further integrate the graph-based technique to get more effective models, which we will discussed in \S \ref{graph-mlml}.

The low-rank assumption is rather flexible and may be exploited in various ways. For example, COCO \cite{DBLP:journals/corr/abs-1805-09156} considers a more complex setting that the features and labels are missing simultaneously. It imposes the concatenation of recovered feature matrix and label matrix to be low-rank via trace norm. Some works also utilize the assumption through a low-rank label correlation matrix. ML-LRC \cite{DBLP:conf/icdm/XuWSWC14} assumes the label matrix can be reconstructed using a correlation matrix $U$ such that $Y=\hat{Y}^TU$, where $U\in\mathbb{R}^{L\times L}$ is low-rank. Then, the loss is measured using the output and the reconstructed labels $||XW-YU||_F^2$. Based on this assumption, ML-LEML \cite{DBLP:journals/kbs/MaTZC17} further involves an instance-wise label correlation matrix $V$ such that $Y=\hat{Y}V$, where $V\in\mathbb{R}^{n\times n}$ is also low-rank.

Another popular way is to follow the paradigm of embedding methods that projects the label vectors to a low-dimensional space. \cite{DBLP:conf/aaai/YehWKW17} proposes a deep neural network-based model C2AE. The features and labels are jointly embedded to a latent space using two neural networks $F_x$ and $F_e$ such that their codewords are maximally correlated. Then, the feature codewords are decoded by another neural network $F_d$, which is also used for prediction. For MLML problems, the decoded labels are evaluated on the observed entries. Though the labels are decoded from a low-dimensional space, the low-rank assumption need not be satisfied with non-linear projection. Thus, REFDHF \cite{DBLP:conf/icmcs/Wang19} added a trace norm regularization term on the decoded label matrix. \cite{DBLP:conf/icmcs/Wang19} also proposes a novel hypergraph fusion technology that explores and utilizes the complementary between feature space and label space. Compared to low-rank classifier-based methods, the embedding methods are more flexible since the classifier can be non-linear and thus are worthy to be explored.

\subsubsection{Graph-based Methods}\label{graph-mlml}
To handle missing labels, one of the most popular solutions is graph-based model. Denote a weighted graph by $\mathcal{G}=(\mathcal{V},\mathcal{E},\mathcal{W})$, where $\mathcal{V}=\{x_i|1\le i\le n\}$ denotes the vertex set and $\mathcal{E}=\{(x_i,x_j)\}$ denotes the edge set. $\mathcal{W}=[w_{ij}]_{n\times n}$ is a weight matrix where $w_{ij}=0$ if $(x_i,x_j)\notin E$. With the graph being defined, the most typical strategy is adding a manifold regularization term to the empirical risk minimization framework. Note that in this section, we slightly abuse the notation $w_{ij}$ to represent the graph weight entry for the sake of simplicity.

The pioneering work \cite{DBLP:conf/aaai/SunZZ10} is the first to propose the concept of multi-label learning with weak labels, i.e. the implicit setting of MLML. The proposed method, named WELL, constructs a label-specific graph for each label from a feature-induced similarity graph. Then, the manifold regularization terms are added separately for each label. \cite{DBLP:conf/icpr/WuLWHJ14} formalized the other setting of MLML and involves three assumptions into MLML according to \cite{DBLP:conf/sdm/ChenSWZ08},
\begin{itemize}
\item \textbf{Label Consistency}: the predicted labels should be consistent with the initial labels, which is usually achieved by empirical risk minimization principle;
\item \textbf{Sample-level Smoothness}: if two samples $x_i$ and $x_j$ are close to each other, so are their predicted label vectors;
\item \textbf{Label-level Smoothness}: if two incomplete label vectors $y_i$ and $y_j$ are semantically similar, so are the predicted label vectors.
\end{itemize}
Formally, a $k$-nearest neighbor graph is constructed to satisfy the sample-level smoothness, where weight matrix $\mathcal{W}^x$ is computed by $w_{ij}^x=\exp(-\frac{||x_i-x_j||_2^2}{||x_i-x_h||_2||x_j-x_h||_2})$, where $x_h$ is the $h$-th nearest neighbor of $x_i$ ($h$ is a fixed constant). For the label-level smoothness, the authors constructs a $L$-square weight matrix $\mathcal{W}^y$ where $w_{ij}^y=\exp(-\eta[1-\frac{\langle \tilde{Y}_{i.},\tilde{Y}_{j.}\rangle}{||\tilde{Y}_{i.}||_2||\tilde{Y}_{j.}||_2}])$. $\tilde{Y}_{i.}$ is the $i$-th row vector of incomplete matrix $\tilde{Y}$. Finally, the predicted score matrix $\dot{Y}$ is learned by,
\begin{equation}\label{MLML}
\begin{split}
\min_{\dot{Y}} ||\dot{Y}-\tilde{Y}||_F^2+\frac{\lambda_x}{2}\textrm{Tr}(\dot{Y}L_x\dot{Y}^T)+\frac{\lambda_y}{2}\textrm{Tr}(\dot{Y}^T L_y\dot{Y})
\end{split}
\end{equation}
where $L_x$ and $L_y$ is the laplacian matrix of $\mathcal{W}^x$ and $\mathcal{W}^y$. $\lambda_x$ and $\lambda_y$ are trade-off parameters. This learning paradigm is followed by some recent works. \cite{DBLP:journals/pr/WuLHJ15} proposes an inductive version that the trained classifier can also predict on unseen data. \cite{DBLP:journals/pr/LiuWGGN18} chooses the hinge loss as the empirical risk instead of squared loss. To tackle the severe class imbalance problem in MLML, \cite{DBLP:conf/aaai/WuLG16} add two class cardinality constraints to Eq. (\ref{MLML}) that enforces the number of positive labels is in a predefined range. With hierarchical label information being provided, MLMG-GO \cite{DBLP:journals/ijcv/WuJLGL18} involves a semantic hierarchical constraints such that the score of a label $y_a$ is smaller than its parent label $y_b$. In \cite{huynh2020interactive}, a new regularization framework IMCL is proposed that interactively learns the two similarity graphs.

Many graph-based methods only concentrate on the sample-level smoothness principle. That is, the graph information is mainly used for disambiguating the incomplete supervision, and different techniques are involved to utilize the label correlations. \cite{DBLP:conf/aaai/YuHDL17} treats the problem of one-class matrix factorization with side information as an MLML task. Inspired by \cite{DBLP:conf/icml/Yu0KD14}, the linear classifier is restricted to be low-rank and the predicted label matrix is smoothed by a manifold regularization term. Since the low-rank assumption fails in many applications, MLMG-SL \cite{DBLP:journals/ijcv/WuJLGL18} further assumes that the output of graph model can be decomposed to a low-rank matrix and a sparse matrix. There are also several recent works that focus on the label-level smoothness. LSML \cite{DBLP:conf/bigmm/HuangQZCYZ18} proposes to learn a label correlation matrix, i.e. a label graph, that can be used to complement the missing labels, smooth the label prediction and guide the learning of label-specific features simultaneously. GLOCAL \cite{DBLP:journals/tkde/ZhuKZ18} trains a low-rank model with manifold regularization that exploits global label-level smoothness. In addition, as label correlations may vary from one local region to another, GLOCAL partitions the instances to several groups and learns local label correlations by group-wise manifold regularization. In \cite{DBLP:conf/cvpr/DurandMM19}, a fully connected graph is built whose vertices are the labels and then, a graph neural network (GNN) is trained to model the label dependencies. The input of GNN is the $L$-sized feature vector of the image extracted by a convolutional neural network, and the outputs are the predicted labels. To disambiguate the missing labels, \cite{DBLP:conf/cvpr/DurandMM19} proposes two novel strategies. For the known labels, the authors propose partial binary cross-entropy loss (Partial-BCE) that reduces the normalization factor according to the label proportion. To complete the missing entries, \cite{DBLP:conf/cvpr/DurandMM19} adopts a curriculum learning strategy that learns a self-paced model.

Besides, some studies are also interested in different graph information. APG-Graph \cite{DBLP:conf/eccv/YangZC16} proposes a novel semantic descriptor-based approach for visual tasks to construct an instance-instance correlation graph. Specifically, \cite{DBLP:conf/eccv/YangZC16} makes use of the posterior probabilities of the classifications on other public large-scale data sets. Then, a $k$NN graph is constructed by these predicted tags. \cite{DBLP:conf/aaai/YuHDL17} regards the user-item interaction in the recommender system as a bipartite graph. 

In the past few years, graph mining techniques have received huge attention. We believe the future graph-based MLML models will involve more expressive graph models, e.g. graph neural networks \cite{DBLP:journals/corr/abs-1812-08434}, and various types of graphs, e.g. social networks \cite{DBLP:conf/naacl/DongWHC19}.


\subsubsection{Other Techniques for Missing Labels}
There are many other techniques can be used for MLML tasks, such as co-regularized learning \cite{DBLP:conf/icml/ChenZW13}, binary coding embedding \cite{DBLP:conf/eccv/WangSWLS14}. In what follows, we focus on some advanced MLML algorithms.

Due to the capability of exploring the data distribution, probability graphical models (PGMs) have been popular for MLML problems since we can complement the missing labels in a generative manner. SSC-HDP \cite{DBLP:conf/kdd/QiYZZ11} involves a correspondence hierarchical Dirichlet process (Corr-HDP) that enables the dimension of latent factors to be chosen dynamically. Based on Corr-HDP, SSC-HDP iteratively updates the likelihood $P(y^j|x)$ for an instance $x$ whose $j$-th label is missing, while the likelihood of remaining labels is fixed to $1$. CRBM \cite{DBLP:conf/aistats/0013ZG15} proposes a conditional restricted Boltzmann machine model to capture the high-order label dependence relationships. In specific, a latent layer is added above the labels layer to form a restricted Boltzmann machine, while the features are the conditions. Based on a latent factor model, GenEML \cite{DBLP:conf/icml/JainMR17} proposes a scalable generative model that involves an exposure variable for each missing labels. BMLS \cite{DBLP:conf/aistats/ZhaoRDB18} jointly learns a low-rank embedding of the label matrix and the label co-occurrence matrix using an Poisson-Dirichlet-gamma non-negative factorization method \cite{DBLP:journals/jmlr/ZhouHDC12}. Note that \cite{DBLP:conf/aistats/0013ZG15,DBLP:conf/aistats/ZhaoRDB18} are also capable of incorporating auxiliary label relatedness information, such as Wikipedia.


Reweighting empirical risks is also a common strategy. \cite{DBLP:conf/cvpr/BucakJJ11} notices that in MLML setting, the traditional multi-label ranking error may overestimate the classification error. Hence, a slack variable is introduced to account for the error of ranking an unassigned class before the assigned class. \cite{DBLP:conf/kdd/JainPV16} proposes an unbiased propensity scored variant of nDCG loss and \cite{DBLP:conf/cvpr/DurandMM19} presents Partial-BCE, which we have discussed in previous sections. \cite{DBLP:conf/mir/IbrahimEPR20} assigns a weight factor for each term in binary cross-entropy loss. In particular, the weights of the positive labels are fixed to $1$. The weights of missing entries are set as $P(\hat{y}_c|y)$, i.e. the probability of having a negative label for the $c$-th label given the vector of labels $y$. Specifically, the probability is estimated from the ground-truth label matrix based on label co-occurrences.

Recently, bandit learning-based approaches are also introduced. Specifically, one pioneering work \cite{DBLP:journals/corr/abs-2102-07800} considers the contextual bandits problem in the extreme multi-label learning context. It modifies the inverse gap weighting sampling strategy to select top-$k$ arms, which results in good generalization performance. Besides, this work proposes a tree-based algorithm by grouping similar arms and thus, the model enjoys a poly-logarithm computational cost w.r.t. the number of arms.

\textbf{Remark} To date, graph-based methods and embedding-based methods are still dominant in the MLML context. Though recent works \cite{huynh2020interactive} have tried to involve deep models to promote performance, they mainly involve trivial convolutional networks and autoencoders. It would be promising to design more tailored model architectures for MLML. Other techniques are also worth to be explored. For example, with the success of existing PGM-based MLML methods, we believe that bayesian deep learning (BDL) \cite{DBLP:conf/icml/KhanNTLGS18} can further improve the performance due to its superiority on high-dimensional data and complex uncertainty.



\subsection{Semi-Supervised Multi-Label Classification}
In semi-supervised MLC (SS-MLC) \cite{DBLP:conf/aaai/LiuJY06}, the data set is comprised of two sets: fully labeled data and unlabeled data. Though SS-MLC has a far longer history than MLML, we can regard it as a special case of MLML, i.e. the labels of some instances are totally missing. In fact, similar to MLML, a plenty of SS-MLC algorithms are also based on graph models \cite{DBLP:journals/tkde/KongNZ13,DBLP:conf/ijcai/Wang2020}, and low-rank assumptions \cite{DBLP:journals/tkde/ZhuKZ18,DBLP:conf/pakdd/SunFLL19}. In what follows, we first review some state-of-the-art SS-MLC algorithms and then, discuss a novel learning setting called weakly-supervised MLC.

\subsubsection{State-of-the-art Algorithms}
Graph-based methods are very popular in SS-MLC, which mainly differ in the strategy of utilizing the label-correlation. SLRM \cite{DBLP:conf/cvpr/JingYYN15} enforces the classifier to be low-rank, while a manifold-regularization term is added to ensure the sample-level smoothness. \cite{DBLP:conf/pakdd/SunFLL19} proposes a triple low-rank regularization approach where the graph is dynamically updated using a low-rank feature-recovery matrix. Based on curriculum learning, ML-TLLT \cite{DBLP:conf/aaai/GongTYL16} forces a teacher pair to generate similar curriculums if the corresponding two labels are highly correlated over the labeled examples. CMLP \cite{DBLP:conf/ijcai/Wang2020} makes use of collaboration technique \cite{DBLP:conf/aaai/FengAH19} to design an scalable multi-label propagation method. Specifically, it breaks the predicted label into two parts: 1) its own prediction part; 2) the prediction of others, i.e. collaborative part.

As mentioned above, other techniques may also be used. COIN \cite{DBLP:conf/kdd/ZhanZ17} adapts the well-known co-training strategy to SS-MLC setting. In each co-training round, a dichotomy over the feature space is learned by maximizing the diversity between the two classifiers induced on either dichotomized feature subset. Then, pairwise ranking predictions on unlabeled data are iteratively communicated for model refinement. Based on COIN, \cite{DBLP:conf/icbk/ChuLH19} further proposes an ensemble method to accommodate streamed SS-MLC data. DRML \cite{DBLP:conf/aaai/WangLQS020} designs a dual-classifier domain adaptation network to align the features in a latent space. In order to model label dependencies, DRML generates the final prediction by feeding the outer-product of the dual predicted label vectors to a relation extraction network.

\subsubsection{Weakly-Supervised MLC}
Due to the large output space, even in the SS-MLC problems, collecting precisely labeled data would take extensive efforts and costs. Hence, a new setting called weakly-supervised multi-label classification (WS-MLC) has attracted enormous attention, i.e. there might be fully labeled data, incompletely-labeled data and unlabeled data in the data set simultaneously. In this survey, we follow the definition of WS-MLC in \cite{DBLP:conf/eccv/ChuYW18}. However, weakly-supervised MLC may also have other meanings in the literature. In a broad sense, any noisy supervision can be termed as weakly-supervision. The readers should also be careful about the difference between WS-MLC and multi-label learning with weak labels \cite{DBLP:conf/aaai/SunZZ10,conf/aaai/WangYL20}. The latter sometimes indicates the implicit setting of MLML problems.

Many effective approaches have been developed to deal with WS-MLC problems. For example, WeSed \cite{DBLP:journals/tbd/WuWZYLZZ15} handles the missing labels by a weighted ranking loss and integrates the unlabeled data via a triplet similarity loss. In \cite{DBLP:journals/ijon/TanYYW17}, missing labels are first estimated by a correlation matrix. Then, a linear classifier is trained by minimizing a graph regularized model. SSWL \cite{DBLP:conf/aaai/DongLZ18} proposes a novel dual similarity regularizer $||Y-VYU||$ to characterize both sample-level and label-level smoothness. Here $V$ is the weight matrix of $k$NN graph over training data and $U$ is a trainable variable that represents the label similarity. Moreover, SSWL also utilizes an ensemble of multiple models to improve the robustness. Though these works have demonstrated promising results, they directly use logical labels, and thus, ignore the relative importance of each label to an instance. To bridge this gap, WSMLLE \cite{DBLP:conf/ijcai/LvXZG19} transforms the original problem to a label distribution learning problem \cite{DBLP:journals/tkde/Geng16}. In specific, a new label enhancement method is proposed that marries the concept of local correlation \cite{DBLP:journals/tkde/ZhuKZ18} and dual similarity regularizer \cite{DBLP:conf/aaai/DongLZ18}. The label enhancement technique is also adopted by fully-supervised MLC \cite{DBLP:conf/icdm/ShaoXG18} and PML \cite{DBLP:conf/aaai/0009LG20} models, and we will give a detailed discussion about the latter one in \S \ref{pml}.

Probabilistic models are also popular in solving WS-MLC tasks, since the distribution of unlabeled data can be seamlessly integrated into a probabilistic framework. DSGM \cite{DBLP:conf/eccv/ChuYW18} proposes a deep sequential generative model which assumes an instance $x$ is generated from its label $y$ as well as a latent variable $z$.
DSGM leverages information from observed labels in a sequential manner. Then, the model is trained by maximizing the likelihood,
\begin{equation}\label{REML}
\begin{split}
\max_\theta\!\!\sum_{i\in D_l}\!\log p_\theta(x_i,y_i)\!+\!\!\sum_{j\in D_o}\!\log p_\theta(x_j,\tilde{y}_j)\!+\!\!\sum_{k\in D_u}\!\log p_\theta(x_k)
\end{split}
\end{equation}
where $\theta$ is the model parameter. $D_l$, $D_o$ and $D_u$ are the index sets of fully labeled data, incompletely-labeled data and unlabeled data respectively. \cite{DBLP:conf/eccv/ChuYW18} also proposes a variational inference method that minimizes the evidence lower bound of the objective. \cite{DBLP:journals/tkde/AkbarnejadB19} designs an embedding-based probability model called ESMC, which addresses some key issues in WS-MLC tasks. Since the low-rank assumption may be broken by tail labels, ESMC uses the gaussian processes to perform non-linear projection. To handle missing labels, ESMC introduces a set of auxiliary random variable, a.k.a. experts, to model the relationship between the real-valued probability score and the observed logical labels. Finally, the unlabeled data can also be integrated to learn a smooth mapping from the feature space to the label space.

\textbf{Remark.} Compared to MLML problems, the presence of a large amount of unlabeled data in SS-MLC can highly restrict the representation ability of the model. However, few efforts have been made to apply tricks in state-of-the-art deep semi-supervised learning to SS-MLC. We recommend involving techniques such as consistency regularization \cite{DBLP:conf/NeurIPS/TarvainenV17} and self-supervised pretraining \cite{DBLP:conf/icml/ChenK0H20}, which have demonstrated exciting ability to utilize the unlabeled data.


\subsection{Partial Multi-Label Learning}\label{pml}
In practice, the complicated structure of the label space usually makes it hard to decide some \textit{hard} labels are relevant or not. For example, it is usually hard to decide whether a dog is a malamute or a husky. One might naively drop these labels and regard the original problem as an MLML task. However, missing labels provides no information to the user at all. Hence, partial multi-label learning (PML) \cite{DBLP:conf/aaai/XieH18} is proposed to address this issue, which preserves all the potentially correct labels. Formally speaking, each instance $x$ is equipped with a set of candidate labels $S$, only some of which are the true relevant labels. The remaining labels are called \textit{false positive labels} or \textit{distractor labels}.
Technically, PML can be regarded as a dual problem of MLML and solved by existing MLML techniques. However, it is worth noting that this strategy may be less practical owing to the sparsity of the label space.
Moreover, PML also provides a safe way to protect data privacy since no label can be determined as the ground-truth, as opposite to MLML data.

\subsubsection{Two-stage Learning Methods}
In PML, while label correlation still matters, the other key issue becomes identifying the ground-truth from the candidate label set instead of completion. To handle these issues, some PML algorithms adopt a two-stage learning framework. Formally, an enriched label representation $\Lambda=[\lambda_{ij}]\in \mathbb{R}^{L\times n}$ will be learned where $\lambda_{ij}$ is a real-valued number. The sign of $\lambda_{ij}$ indicates whether the label is positive or negative, while the magnitude reflects the confidence of the relevance. Then, the PML problem is transformed into a canonical supervised learning problem and the classifier can be easily induced. To obtain $\Lambda$, PARTICLE \cite{PARTICLE9057438} uses the label propagation technique that aggregates the information from the $k$-nearest neighbors. After that, the confidences are converted back to logical labels by thresholding. To train an MLC classifier, \cite{PARTICLE9057438} adopts a pairwise label ranking model coupled with virtual label splitting or maximum a posteriori (MAP) reasoning. PARTICLE has two main drawbacks. First, the confidences have richer information than logical labels, but, it is trimmed when thresholding. Second, only the second-order label correlation is considered.

To tackle these problems, DRAMA \cite{DBLP:conf/ijcai/Wang0ZZHC19} generates the label confidence matrix under the guidance of feature manifold and the candidate label set. Then, a novel gradient boosting decision tree (GBDT) based multi-output regressor is directly trained on the transformed data set $\tilde{\mathcal{D}}=\{(x_i,\lambda_i)|i\in\{1,\ldots,n\}\}$ where $\lambda_i$ is the $i$-th column vector of $\Lambda$. On $t$-th boosting round, DRAMA augments the feature space using previously learned labels. Therefore, high-order label correlations are automatically exploited to improve performance.

The major limitation of the aforementioned methods is that the disambiguation is achieved purely by features. However, label correlation itself can help to identify the correct labels. Insufficient disambiguation makes the induced MLC classifier error-prone. To this end, PML-LD \cite{DBLP:conf/aaai/0009LG20} proposes a novel label enhancement method that transforms the PML problem to a label distribution learning problem  \cite{DBLP:journals/tkde/Geng16}. When learning the label confidence matrix, PML-LD leverages the sample-level smoothness and local label-level smoothness \cite{DBLP:journals/tkde/ZhuKZ18} such that the candidate label set can be fully disambiguated. Then, the confidences are normalized by softmax to form an LDL problem and a multi-output support vector machine is induced.


The advantages of two-stage PML methods are two-folds. First, since the label confidences are obtained, we can apply well-studied multi-output learning methods \cite{DBLP:journals/corr/abs-1901-00248}. Second, the real-valued confidences reflect the relative intensity of the relevance or irrelevance, which may give us more information about our data.

\subsubsection{End-to-end Learning Methods}
As we have mentioned, two-stage learning PML methods usually need be carefully designed, or the induced MLC classifier may be error-prone due to insufficient disambiguation. Hence, many PML algorithms are developed in an end-to-end fashion, which vary from one to another.

\cite{DBLP:conf/aaai/XieH18} proposes a ranking model, which employs the label confidence as a weight for the ranking loss. To estimate the label confidences, \cite{DBLP:conf/aaai/XieH18} provides two practical ways based on label correlation and feature prototypes respectively. Moreover, the classification model along with the ground-truth confidence are optimized in a unified framework such that the two subproblems can benefit from each other. \cite{DBLP:conf/icdm/HeD0SL19} presents a soft sign thresholding method to measure the discrepancy between the real-valued confidences and the candidate labels. Similar to \cite{DBLP:conf/aaai/XieH18}, the classifier training and disagreement minimization are performed at the same time. Nevertheless, \cite{DBLP:conf/icdm/HeD0SL19} does not well utilize the label correlations, and thus the performance is limited.

Some methods adopt the low-rank assumption. fPML \cite{DBLP:conf/icdm/YuCDWLZW18} introduces the matrix factorization technique to obtain a shared latent space for both features and labels. The classifier is then trained by fitting the recovered labels. PML-LRS \cite{DBLP:conf/aaai/SunFWLJ19} utilizes the low-rank and sparse decomposition scheme. That is, it assumes the distractor label matrix is sparse while the ground-truth matrix is low-rank. Both fPML and PML-LRS treat the false-positive labels as randomly generated noise. However, in real-world applications, the false-positive labels may be caused by some ambiguous contents of the instance. Therefore, \cite{DBLP:conf/aaai/XieH20} divides the classifier $W$ to two parts $W=U+V$. Here $U$ is the multi-label classifier and $V$ is the distractor label identifier. Meanwhile, $U$ is constrained to be low-rank to utilize label correlations. Since distractor labels usually correlate to only a few ambiguous features, $V$ is regularized to be sparse. MUSER \cite{DBLP:conf/ijcai/LiLF20} takes redundant labels together with noisy features into account by jointly exploring feature and label subspaces. Furthermore, it uses a manifold regularizer to ensure the consistency between features and latent labels.

\textbf{Remark.} The PML problem is drawing increasing attention in the community. However, the assumption that all labels are equally being candidates can be less practical, since some ground-truth labels can be easily distinguished. Therefore, the key assumptions of PML should be carefully revisited. Here we recommend a more practical setting that besides providing the candidate set, the annotators should also provide partial ranks that which labels are more likely to be correct. Besides, existing PML data sets are mainly built upon multi-instance multi-label \cite{DBLP:journals/isci/HeGW12} data sets, and thus, there is also an urgent need to establish a benchmark for PML problems.

\subsection{Other Settings}
The complexity of the label space has expedited various kinds of improperly-supervised MLC settings. In what follows, we briefly review some more state-of-the-art settings in the literature.

\textbf{MLC with Noisy Labels (Noisy-MLC).} While MLML and PML consider single-side noise, Noisy-MLC assumes that noisy labels occur in both relevant and irrelevant labels. Many effective Noisy-MLC algorithms have been proposed to address this problem, including graph based methods \cite{DBLP:journals/tcyb/ZhangYFZCH20}, probability models \cite{DBLP:conf/aaai/CuiZJ20}, teacher-student model \cite{DBLP:conf/accv/HuHSC18}. In \cite{DBLP:conf/ijcai/LvXZG19}, the WS-MLC framework is extended and noisy labels are assumed to be contained in the data set. Some works \cite{DBLP:conf/cvpr/VeitACKGB17,DBLP:conf/cvpr/HuHSC19} maintain a small set of clean data to reduce the noise in the large data set. Since learning from label noise have been a hot topic in the community, Noisy-MLC deserve more attention.

\textbf{MLC with Unseen Labels.} In the aforementioned settings, the label spaces is fixed during training and testing. However, in practice, the label space may be dynamically expanded. For instance, \cite{DBLP:journals/tkde/ZhuTZ18} studies an online MLC setting that an arriving data instance may be associated with unknown labels.
In \cite{Wang2020}, knowledge distillation method is used to handle streaming labels. Multi-label zero-shot learning (ML-ZSL) \cite{DBLP:conf/cvpr/LeeFYW18} requires the prediction of unknown labels which are not defined during training. To make ML-ZSL feasible, external semantic information is usually involved, such as word vectors \cite{DBLP:conf/cvpr/ZhangGS16} and knowledge graphs \cite{DBLP:conf/cvpr/LeeFYW18}. In \cite{DBLP:conf/cvpr/AlfassyKASHFGB19}, few-shot labels is also considered, which relates to only few instances in the data set, i.e. nearly unseen.

\textbf{Multi-Label Active Learning (MLAL).} Active learning is a notable way to alleviate the difficulty of multi-label tagging. The idea is to carefully select the most informative data instances for labeling such that better models can be trained with less labeling effort. A variety of works have studied MLAL problems. For example, \cite{DBLP:conf/kdd/YangSWC09} adopts maximum loss reduction with
maximal confidence as the sampling criterion for MLAL. \cite{DBLP:conf/icml/Shi019} solves MLAL problems via a probability model. Moreover, MLAL is also considered in crowdsourcing \cite{DBLP:journals/tkde/LiJCZ19} and novel queries \cite{DBLP:conf/ijcai/HuangCZ15} tasks.

\textbf{Label Distribution Learning (LDL).} LDL \cite{DBLP:journals/tkde/Geng16} is a general framework that assigns $L$ normalized real values to label description degree. It aims to tackle inherent ambiguity in data annotation, e.g. a facial expression usually conveys a complex mixture of basic emotions. Since it is difficult to obtain the label distribution directly, many works \cite{DBLP:conf/icdm/ShaoXG18,DBLP:conf/icml/0009SLG20,DBLP:conf/ijcai/XuTG18} focus on recovering label distributions from logical labels, which is also known as label enhancement (LE). LE is an effective learning strategy to deal with label ambiguity. LE is also applied in WS-MLC \cite{DBLP:conf/ijcai/LvXZG19} and PML \cite{DBLP:conf/aaai/0009LG20} to handle imperfect supervision signals.

\textbf{MLC with Multiple Instances (MIML).} MIML \cite{DBLP:journals/isci/HeGW12} is a popular setting which assumes each example is described by multiple instances as well as associated with multiple binary labels. Recent studies in MIML \cite{DBLP:conf/cvpr/YangZCO17,DBLP:conf/aaai/ZhangSLL20} have developed many deep learning models such that noisy instances can be effectively figured out. Nevertheless, MIML mainly focuses on the instance-level ambiguity instead of the labels. Hence, we do not further discuss it.

\textbf{Remark.} Intelligent systems are enrolled in increasingly difficult and complicated tasks, and thus new settings like PML and LDL deserve more attention. Moreover, there remain more challenging and complicated settings in real-world applications to be explored. For instance, there might be out-of-distribution detection \cite{DBLP:conf/iclr/HendrycksG17}, domain shift \cite{DBLP:journals/ijon/WangD18} and other problems arise in MLC problems.

\section{Deep Learning for Multi-label Learning}\label{MLCwithDL}

Due to the powerful learning capability, deep learning has achieved state-of-the-art performance in many real-world multi-label applications, e.g., multi-label image classification.
In MLC problems, it is key to harvest the advantage of deep learning to better capture the label dependencies.
In this section, we first introduce some representative deep embedding methods for MLC, then present deep learning for challenging MLC, and finally review advanced deep learning for MLC.

\subsection{Deep Embedding Methods for MLC}

Different from conventional multi-label methods, deep neural networks (Deep NNs) often seek a new feature space and employ a multi-label classifier on the top.
To our knowledge, BP-MLL \cite{Zhang:2006:MNN:1159162.1159294} is the first method to utilize NN architecture for multi-label learning problem.
To explicitly exploit the dependencies among labels, given the neural network $F$, BP-MLL introduces a pairwise loss function for each instance $x_i$:
\begin{equation} \label{E:BP}
  {E}_i =  \frac{1}{\lvert y_i^1 \rvert  \lvert y_i^0 \rvert }  \sum_{(p,q) \in y_i^1 \times y_i^0} \exp \left( -( F(x_i)^p - F(x_i)^q )  \right)
\end{equation}
where $y_i^1$ and $y_i^0$ denote the sets of positive and negative labels for the $i$-the instance $x_i$ respectively, $(F(x_i))^p$ denotes the $p$-th entry of $F(x_i)$.
$F(x_i))^p - (F(x_i))^q$ measures the difference between the outputs of the network on the positive and negative labels, and the exponential function is used to severely penalize the difference.
Thus the minimization of \eqref{E:BP} leads to output larger values for positive labels, and smaller values for the negative labels.
\cite{Zhang:2006:MNN:1159162.1159294} further shows that \eqref{E:BP} is closely related to \textit{ranking loss}.

Later, \cite{DBLP:conf/pkdd/NamKMGF14} finds that BP-MLL does not perform as expected on data sets in textual domain.
To address the issue, based on BP-MLL, \cite{DBLP:conf/pkdd/NamKMGF14} proposes to use a comparably simple NN approach that can achieve the state-of-the-art performance in large-scale multi-label text classification.
They show that the ranking loss in BP-MLL can be efficiently and effectively replaced by the commonly used cross-entropy function, and several NN tricks, i.e., rectified linear units (ReLUs), Dropout, and AdaGrad can be effectively employed in this setting.

Embedding methods have been effective to capture the label dependency and reduce the computation costs.
However, existing embedding methods are shallow models, which may not be powerful to discover high order dependency among labels.
To fulfill this gap, \cite{DBLP:conf/aaai/YehWKW17} proposes Canonical Correlated AutoEncoder (C2AE), which is the first DNN-based embedding method for MLC to our knowledge.
The basic idea of C2AE is to seek a deep latent space to jointly embed the instances and labels.
C2AE performs feature-aware label embedding and label-correlation aware prediction. The former is realized by joint learning of deep canonical correlation analysis (DCCA) and the encoding stage of autoencoder, while the latter is achieved by the introduced loss function for the decoding outputs.

C2AE consists of two DNN modules, i.e., DCCA and autoencoder, and seeks three mapping functions, i.e., feature mapping $F_x$, encoding function $F_e$, and decoding function $F_d$.
For training, C2AE receives instance $X$ and labels $Y$,  associates them in the latent space $L$, and enforces the recover of $Y$ using autoencoder.
The objective function of C2AE is defined as follows:
\begin{equation}
  \min_{F_x, F_e, F_d} \Phi(F_x, F_e) + \alpha \Gamma(F_e, F_d)
\end{equation}
where $\Phi(F_x, F_e)$ and $\Gamma(F_e, F_d)$ denote the losses in the latent and output spaces respectively, $\alpha$ is used to balance the two terms.
Inspired by the CCA, C2AE learns the deep latent space by maximizing the correlation between instances and labels. Thus $\Phi(F_x, F_e)$ can be defined as:
\begin{alignat}{2}
\min_{F_x, F_e} & \quad \| F_x(X) - F_e(Y)  \|_F^2 \\
\textrm{s.t.} & \quad F_x(X) F_x(X)^T = F_e(Y) F_e(Y)^T = I \notag
\end{alignat}
In addition, C2AE recovers the labels using autoencoder with aim of preserving label dependency.
Inspired by \cite{Zhang:2006:MNN:1159162.1159294}, $\Gamma(F_e, F_d)$ is defined as follows:
\begin{alignat}{2} \label{E:C2AE}
  & \Gamma(F_e, F_d)  \!\!=\!\!  \sum_{i=1}^N E_i  \\
  & E_i \!\! =\!\! \frac{1}{\lvert y_i^1 \rvert  \lvert y_i^0 \rvert }  \!\!\!\sum_{(p,q) \in y_i^1 \times y_i^0}\!\!\! \exp \left( -( \!F_d(F_e(x_i))^p\!\! - \!\!F_d(F_e(x_i))^q\! )  \right) \notag
\end{alignat}
where $N$ is the number of the instances, $F_d(F_e(x_i))$ is the recovered label of $x_i$ using the autoencoder.
For prediction, given a test instance $\hat{x}$, C2AE performs prediction as $\hat{y} = F_d(F_x(\hat{x}))$.

Later, inspired by C2AE, \cite{DBLP:conf/aaai/ChenW0ZH019} presents a two-stage label embedding model based on neural factorization machine model. It first exploits second-order label correlation via a factorization layer and then learns high-order correlation by additional fully-connected layers. \cite{DBLP:conf/acml/WangYYY18} proposes another deep embedding method, i.e., Deep Correlation Structure Preserved Label Space Embedding (DCSPE).
In addition to DCCA, DCSPE further develops deep multidimensional scaling (DMDS) to preserve the intrinsic structure of the latent space.
Finally, DCSPE transforms test instance into the latent space, searches its nearest neighbor, and finally regards label of this neighbor as prediction.
However, as the $k$NN search is time-consuming, the $k$NN embedding methods are computationally expensive in the large-scale setting.
To solve the above issue, \cite{DBLP:conf/ijcai/0001LLOT18} proposes a novel deep binary prototype compression (DBPC) for fast multi-label prediction.
DBPC compresses the database into a small set of short binary prototypes, and uses the
prototypes for prediction.

For multi-label emotion classification, \cite{DBLP:conf/aaai/FeiZRJ20} recently proposes latent emotion memory (LEM) to learn latent emotion distribution without external knowledge.
LEM includes latent emotion and memory modules to learn emotion distribution and emotional features respectively, and the concatenation of the two is fed into Bi-directional Gated Recurrent Unit (BiGRU) for prediction.
For multi-label image classification, \cite{DBLP:conf/cvpr/ZhuLOYW17}  proposes a unified deep neural network that exploits both semantic and spatial relations between labels with only image-level supervision.
Specifically, the authors propose Spatial Regularization Network (SRN) that generates attention maps for all labels and captures the underlying
relations between them via learnable convolutions.
\cite{DBLP:conf/cvpr/0002ZFY019} finds the consistency of attention regions of CNN classifiers under many transforms are not preserved.
To address the issue, the authors propose a two-branch network with original and transformed images as inputs and introduce a new attention consistency loss that measures the attention heatmap consistency between two branches.
Later \cite{DBLP:conf/aaai/YouGCLBW20} proposes Adjacency-based Similarity Graph Embedding (ASGE) and Cross-modality Attention (CMA) to capture the dependencies between labels and discover locations of discriminative features respectively.
Specifically, ASGE learns semantic label embedding that can explicitly exploit label correlations, and  CMA generates the meaningful attention maps by leveraging more prior semantic information.
Instead of requiring laborious object-level annotations, \cite{DBLP:conf/mm/LiuSSYXP18} proposes to distill knowledge from weakly-supervised detection (WSD) task to boost MLC performance.
The authors construct an end-to-end MLC framework augmented by a knowledge distillation module that guides the classification model by the WSD model for object RoIs.
WSD and MLC are the teacher and student models respectively.

\textbf{Remark.}
Deep embedding methods are the most widely-used deep methods for MLC.
Among them, C2AE is pioneer deep embedding work for MLC and has been applied in many real-world applications including multi-label emotion classification, which deserves exploration for the beginners to understand basic mechanism.
As we know, label correlation is key for MLC, and some objectives, e.g., \eqref{E:C2AE} have been used to model label correlation in deep embedding methods for MLC.
However, existing research shows that \eqref{E:C2AE} may not be effective for textual domain.
In the future, how to effectively capture label correlation is an important research topic of deep embedding methods for MLC.
Some advanced techniques, e.g., graph convolutional network (GCN), recurrent neural network (RNN) open doors to better capture label correlation, and can motivate further research of deep embedding methods for MLC.

\subsection{Deep Learning for Challenging MLC}

In real world applications, multi-label learning is often challenging due to the complex setting of labels.
For instance, the number of labels is very large known as XMLC; the labels are often partially or weakly given; labels emerge continuously or are even unseen before.
This section reviews the recent advances of deep learning to address these challenging MLC problems.

\textbf{DL for Extreme MLC.}
To our knowledge, \cite{DBLP:conf/sigir/LiuCWY17} is the first attempt at applying deep learning to XMLC.
XML-CNN \cite{DBLP:conf/sigir/LiuCWY17} applies convolutional neural network (CNN) and dynamic pooling to learn the text representation, and a hidden bottleneck layer much smaller than the output layer is used to achieve computational efficiency.
However, XML-CNN still suffers from effectiveness of capturing the important subtext for
each label.
To address this issue, AttentionXML \cite{DBLP:conf/NeurIPS/YouZWDMZ19} is proposed with two unique features: 1) a multi-label attention mechanism with raw text as
input, which allows to capture the most relevant part of text to each label, 2) a shallow and wide probabilistic label tree (PLT), which allows to handle millions
of labels, especially for "tail labels".
Meanwhile, based on C2AE, a new deep embedding method, i.e., Ranking-based Auto-Encoder (Rank-AE) \cite{DBLP:conf/naacl/WangCSQLZ19} is proposed for XMLC.
Rank-AE first uses an efficient attention mechanism to learn rich representations from any type of input features, learns a latent space for instance and labels, and finally develops a margin-based ranking loss that is more effective for XMLC and noisy labels.
\cite{DBLP:conf/NeurIPS/GuoMWHKRK19} empirically demonstrates that overfitting leads to the poor performance of the DNN based embedding methods for XMLC.
Based on this finding, \cite{DBLP:conf/NeurIPS/GuoMWHKRK19} further proposes a new regularizer, i.e., GLaS for embedding-based neural network approaches.
\cite{DBLP:conf/kdd/ChangYZYD20} finetunes a pretrained deep transformer for better feature representation. They propose a novel label clustering model for XMLC and the transformer serves as a neural matcher.
With the proposed techniques, the state-of-the-art performance is achieved on several widely-used extreme data sets.
Very recently \cite{Dahiya2021} develops DeepXML framework that can generate a family of algorithms by including four sub-tasks, i.e., intermediate representation, negative sampling, transfer learning, and classifier learning.
It yields Accelerated Short Text Extreme Classifier (Astec) that is more accurate and faster than state-of-the-art deepXMLs on public short text data sets.
DECAF \cite{Mittal2021} considers label metadata, e.g., textual descriptions of labels, which is informative but usually ignored by existing methods.
DECAF jointly learns model enriched by label metadata and feature representation, and predicts accurately with millions of labels.
ECLARE \cite{Mittal2021WWW} incorporates label text and label correlations, and develops frugal architecture and scalable techniques to train model with label correlation graph with millions of labels.
Similarly, GalaXC \cite{Saini2021WWW} collaboratively learns over joint document-label graphs that can incorporate various sources, e.g., label metedata.
GalaXC further introduces label-wise attention to obtain high-capacity extreme classifiers.
\cite{Saini2021WWW} shows that GalaXC is up to 18\% more accurate than state-of-the-arts while it trains 2-50 times faster and predicts 10 times faster on benchmark data sets.

\textbf{DL for partial and weakly-supervised MLC.}
Several efforts \cite{9529072,DBLP:conf/cvpr/DurandMM19,huynh2020interactive,9447152} have been made towards MLC with partial labels.
\cite{DBLP:conf/cvpr/DurandMM19} empirically shows that  partially annotating all images is better than fully annotating a small subset.
Thus \cite{DBLP:conf/cvpr/DurandMM19} generalizes the standard binary cross-entropy loss by exploiting label proportion information, and develops an approach based on Graph Neural Networks (GNNs) to explicitly model the correlation between categories.
Later, \cite{huynh2020interactive} regularizes the cross-entropy loss with a cost function that measures the smoothness of labels and features of images on data manifold, and develops an efficient interactive learning framework where similarity learning and CNN training interact and improve each another.
\cite{DBLP:conf/eccv/ChuYW18} is the first deep generative model to tackle weakly-supervised MLC (WS-MLC).
\cite{DBLP:conf/eccv/ChuYW18} proposes a probabilistic framework that integrates sequential prediction and generation processes to exploit information from unlabeled or partially labeled data.

\textbf{DL for MLC with unseen labels.}
In conventional MLC, all the labels are assumed to be fixed and static; however, it is ignored that labels emerge continuously in changing environments.
To fulfill this gap, a novel DNN-based method, i.e., Deep Streaming Label Learning (DSLL) \cite{Wang2020} is proposed to deal with MLC with newly emerged labels effectively.
DSLL uses streaming label mapping, streaming feature
distillation, and senior student network to explore the knowledge from past labels and historical models to understand new labels.
In addition, \cite{Wang2020} further theoretically proves that DSLL admits tight generalization error bounds for new labels in the DNN framework.
Different from DSLL, \cite{DBLP:conf/cvpr/LeeFYW18} incorporates the additional knowledge graphs for  multi-label zero-shot learning (ML-ZSL).
\cite{DBLP:conf/cvpr/LeeFYW18} advances label propagation mechanism in the semantic space, enabling the reasoning of the learned model for predicting unseen labels.

\textbf{Remark.}
MLC problem is challenging due to the high complexity of labels.
\cite{Saini2021WWW}, \cite{DBLP:conf/cvpr/DurandMM19}, \cite{DBLP:conf/eccv/ChuYW18}, and \cite{Wang2020} are representative deep works for beginners to address extreme MLC, partial MLC,  weakly-supervised MLC, and MLC with unseen labels, respectively.
The above attempts only focus on challenges of label space in MLC problem.
However, in real-world MLC problems, there are some challenges in feature space, e.g., some features may be vanished or augmented, the distribution may change.
How to simultaneously address challenges in label and feature spaces is more challenging and can be regarded as future research of challenging MLC problem.

\subsection{Advanced Deep Learning for MLC}
Recently some advanced deep learning architectures have been developed for MLC problems.

To exploit the underlying rich label structure, \cite{DBLP:conf/icml/CisseAB16} proposes Deep In the Output Space (ADIOS) to partition the labels into a Markov Blanket Chain and then apply a novel deep architecture that exploits the partition.
In multi-label image classification, CNN-RNN \cite{DBLP:conf/cvpr/WangYMHHX16} utilizes recurrent neural networks (RNNs) to better exploit the higher-order label dependencies of an image.
CNN-RNN learns a joint image-label embedding to characterize the semantic label dependency as well as the image-label relevance, and it can be trained end-to-end from scratch to integrate both information in a unified framework.
In addition, instead of using classifier chain, \cite{DBLP:conf/NeurIPS/NamMKF17} proposes to use RNN to convert MLC into a sequential prediction problem,  where the labels are first ordered in an arbitrary fashion.
The key advantage is to allow focusing on the prediction of only positive labels, a much smaller set than the full set of possible labels.
\cite{DBLP:conf/iccv/WangCLXL17} employs Long-Short Term Memory (LSTM) sub-network to sequentially predict semantic labeling scores on the located regions and capture the global dependencies of these
regions, and achieve superior performance in large-scale multi-label image classification.
\cite{DBLP:conf/aaai/ChenCYW18} does not require pre-defined label orders.
It integrates and learns visual attention and LSTM layers for multi-label image classification.
Instead of a fixed, static label ordering, \cite{DBLP:conf/icml/NamKMPSF19} assumes a dynamic, context-dependent label ordering.
\cite{DBLP:conf/icml/NamKMPSF19} consists of a simple EM-like algorithm that bootstraps the learned model, and a more principled approach based on reinforcement learning.
The experiments empirically show dynamic label ordering approach based on reinforcement learning outperforms RNN with fixed label ordering.
\cite{DBLP:conf/aaai/TsaiL20} proposes a new framework based on optimal completion distillation and multitask learning that also does not require a predefined label order.
Recently \cite{DBLP:conf/cvpr/YaziciGRT020} proposes predicted label alignment (PLA) and minimal loss alignment (MLA) to dynamically order the ground truth labels with the predicted label sequence.
This allows for faster training of more optimal LSTM models, and obtains state-of-the-art results in large-scale image classification.

Graph Convolutional Network (GCN) \cite{DBLP:journals/corr/abs-1812-08434} has been also used to successfully model label dependency in MLC problem.
In multi-label image classification problem, \cite{DBLP:conf/cvpr/ChenWWG19} first builds a directed graph over the object labels, employs GCN to model the correlations between labels, and maps label representation to inter-dependent object classifiers.
Similarly, Semantic-Specific Graph Representation Learning (SSGRL) \cite{DBLP:conf/iccv/ChenXHWL19} includes semantic decoupling and interaction modules to learn and  correlate semantic-specific representations respectively.
The correlation is achieved by GCN on a graph built on label co-occurrence.
Later, \cite{DBLP:conf/aaai/WangHLLZMW20} adds lateral connections between GCN and CNN at shallow, middle and deep layers such that label information can be better injected into backbone CNN for label awareness.
For multi-label patent classification, which is regarded as multi-label text classification  problem, \cite{DBLP:conf/aaai/Tang0XPWC20} proposes a new deep learning model based on GCN to capture rich semantic information.
The authors design an adaptive non-local second-order attention layer to model long-range semantic dependencies in text content as label attention for patent categories.

As an alternative of DNN, deep forest \cite{DBLP:conf/ijcai/ZhouF17} is a recent deep learning framework based on tree model ensembles, which does not rely on backpropagation.
\cite{DBLP:journals/corr/abs-1911-06557} introduces deep forest for MLC due to the advantages of deep forest models.
The proposed Multi-Label Deep Forest (MLDF) can handle two challenging problems in MLC: optimizing different performance measures and reducing overfitting.
The extensive experiments show that MLDF achieves the best performance over hamming loss, one-error, coverage, ranking loss, average precision and macro-AUC measures.

\textbf{Remark.}
Advanced deep architecture has more powerful learning capability, and thus can be more effective for MLC problem.
Beginners can try representative deep works of advanced RNN \cite{DBLP:conf/cvpr/WangYMHHX16}, GCN \cite{DBLP:conf/cvpr/ChenWWG19} \cite{DBLP:journals/corr/abs-1911-06557} to address MLC problems.
However, these advanced deep methods usually contain very large amounts of parameters, and require high complexity in terms of training and prediction costs.
To devise lightweight architecture for efficient training and prediction is worthy to be explored for advanced deep methods for MLC.

\section{Online Multi-label Learning}\label{OnlineMLC}
Many real-world applications generate a massive volume of streaming data. For example, many web-related applications, such as Twitter and Facebook posts and RSS feeds, are attached with multiple essential forms of categorization tags. In the search industry, revenue comes from clicks on ads embedded in the result pages. Ad selection and placement can be significantly improved if ads are tagged correctly. This scenario, referred to as online multi-label learning, is a popular tool for addressing large-scale multi-label classification tasks.

The current off-line MLC methods assume that all data are available in advance for learning. However, there are two major limitations of designing MLC methods under such an assumption: firstly, these methods are impractical for large-scale data sets, since they require all data sets to be stored in memory; secondly, it is non-trivial to adapt off-line multi-label methods to the sequential data. In practice, data is collected sequentially, and data that is collected earlier in this process may expire as time passes. Therefore, it is non-trivial to propose new online multi-label learning
methods to deal with streaming data. This section presents a review of the latest algorithms on online multi-label classification.

\cite{DBLP:conf/smc/ErVW16} proposes an online universal classifier (OUC) to handle binary, multi-class and multi-label classification problems. To adapt all types of classification, OUC pre-processes the data set that the target label of all three classification types is represented as a vector with dimension equal to the number of output labels. A deep learning model is then employed for online training.

Based on ELM \cite{DBLP:journals/air/DingZZXN15}, which is a single hidden layer feedforward neural network model, \cite{DBLP:journals/evs/VenkatesanEDPW17} proposes the OSML-ELM approach to handle streaming multi-label data. OSML-ELM uses a sigmoid activation function and outputs weights to predict the labels. In each step, the output weight is learned from the specific equation. OSML-ELM converts the label set from single to multiple representation in order to solve multi-label classification problems.

OLANSGD \cite{DBLP:conf/icassp/ParkC13} is proposed based on label ranking, where the ranking functions are learned by minimizing the ranking loss in the large margin framework. However, the memory and computational costs of this process are expensive on large-scale data sets. Stochastic gradient descent (SGD) approaches update the model parameters using only the gradient information calculated from a single label at each iteration. OLANSGD minimizes the primal form using Nesterov's smoothing, which has recently been extended to the stochastic setting.

However, none of these methods analyze the loss function, and do not use the correlations between labels and features. Some works have been developed to address this issue. For example, \cite{DBLP:journals/ml/ChuHL19} presents a novel cost-sensitive dynamic
principal projection (CS-DPP) method for online MLC. Inspired by matrix stochastic gradient, they develop an efficient online dimension reducer, and provide the theoretical guarantee for their carefully-designed online regression learner. Moreover, the cost information is embedded into label weights to achieve cost-sensitivity along with theoretical guarantees. However, CS-DPP can not capture the joint information between features and labels. To capture such joint information, \cite{DBLP:conf/aaai/GongYB20} proposes a novel online metric learning paradigm for MLC. They first project features and labels into the same embedding space, and then the distance metric is learned by enforcing the constraint that the distance between embedded instance and its correct label must be smaller than the distance between the embedded instance and other labels. Moreover, an efficient optimization algorithm is present for the online MLC. Theoretically, the upper bound of cumulative loss is analyzed in the paper. The experiment results show that their proposed algorithm outperforms the aforementioned baselines.

Recently, some works \cite{DBLP:conf/ijcnn/BoulbazineCMB18,DBLP:conf/ijcai/LiWBS20} study online SS-MLC problem, where data examples can be unlabeled. \cite{DBLP:conf/ijcnn/BoulbazineCMB18} proposes a growing neural gas-based method, which constructs a dynamic graph with incoming data. OnSeML \cite{DBLP:conf/ijcai/LiWBS20} adopts a label embedding fashion that a regression model is learned to fit the latent label vectors. To incorporate the unlabeled data, it extends the regularized moving least-square model \cite{DBLP:conf/iccv/YeungC07} with a local smoothness regularizer. It is noteworthy that online learning from weakly-supervised data has long been a difficult issue since the global data structure is not given \cite{WangECML2020}. Hence, it is valuable to develop online MLC classifiers with limited supervision.

\textbf{Remark.}
Online multi-label learning opens a new way to address large-scale MLC issues with limited memory. Unfortunately, the model, algorithm and theoretical results obtained so far are very limited. It is imperative to put more effort to explore this direction.

\section{Statistical Multi-label Learning}\label{StatisticalMLC}

The generalization error of multi-label learning is analyzed by many papers. For example, \cite{DBLP:conf/icml/Yu0KD14} formulates MLC as the problem of learning a low-rank linear model in the standard ERM framework that could use a variety of loss functions and regularizations. They analyze the generalization error bounds for low-rank promoting trace norm regularization.
There are also some statistical theoretical works which focus on the consistency of multi-label learning: whether the expected loss of a learned
classifier converges to the Bayes loss as the training set size increases. For example, \cite{DBLP:journals/ai/GaoZ13} studies two well-known multi-label loss functions: ranking loss and hamming loss. They provide a sufficient and necessary condition for the consistency of multi-label learning based on surrogate loss functions. For hamming loss, they propose a surrogate loss function which is consistent for the deterministic case. However, no convex surrogate loss is consistent with the ranking loss. \cite{DBLP:conf/icml/DembczynskiKH12} transforms MLC into the bipartite ranking problem, and proposes a simple univariate convex surrogate loss (exponential or logistic) defined on single labels, which is consistent with the ranking loss with explicit regret bounds and convergence rates. Recently, \cite{DBLP:conf/NeurIPS/WydmuchJKBD18} shows that the pick-one-label can not achieve zero regret with respect to the Precision@$k$, and PLTs model can get zero regret (i.e., it is consistent) in terms of marginal probability estimation and Precision@$k$ in the multi-label setting. Inspired by \cite{DBLP:conf/NeurIPS/WydmuchJKBD18}, \cite{DBLP:conf/NeurIPS/X19} further studies the consistency of one-versus-all, pick-all-labels, normalised one-versus-all and normalised pick-all-labels reduction methods based on a different Recall@$k$ metric. All these works study the generalization error and consistency of learning approaches which address multi-label learning by decomposing into a set of binary classification problems. However, the existing theory of the generalization error and consistency does not consider label correlations, and desire for more effort to explore.

As mentioned above, several XMLC methods, such as PD-Sparse \cite{DBLP:conf/icml/YenHRZD16} and SLEEC \cite{DBLP:conf/NeurIPS/BhatiaJKVJ15}, use $\ell_1$ regularization to exploit the sparsity of the data. However, $\ell_1$ regularization suffers two major limitations: 1) \cite{HuiZou2006,ZhangandHuangJ2008,ZhangJ2010} show that the $\ell_1$ regularization produces a bias into the resulting estimator, and harms the estimation accuracy. 2) \cite{JianqingFanandRunzeLiJ2001} proves that the oracle property does not hold for $\ell_1$ regularization. To address these issues, \cite{DBLP:conf/icml/LiuS19} proposes a unified framework for SLEEC with nonconvex penalty, such as minimax concave penalty (MCP) \cite{ZhangJ2010} and smoothly clipped absolute deviation (SCAD) penalty \cite{JianqingFanandRunzeLiJ2001}, which have recently attracted much attention because they can eliminate the estimation bias and attain attractive statistical properties. Theoretically, they show that their proposed estimator enjoys oracle property, which performs as well as if the underlying model were known beforehand, as well as attains a desirable statistical convergence rate of $\mathcal{O}(\frac{\sigma \sqrt{\varpi} +\sqrt{s^*}}{\mu\sqrt{n}} )$, where $\sigma,\varpi, \mu$ are positive constants, $n$ is the sample size and $s^*$ denotes the cardinality of the true support of underlying model. Considering the magnitude of the entries in the underlying model, they can achieve a refined convergence rate of $\mathcal{O}(\frac{\sqrt{s^*}}{\mu\sqrt{n}} )$ under suitable conditions. This paper could inspire the community to bring more powerful statistical penalty method and theory into MLC.

\textbf{Remark.} A key challenging issue in MLC is to model the interdependencies between labels and features. Existing methods, such as classifier chain, CCA and CPLST, attempt to model the correlations between labels and features. However, the statistical properties of these multi-label dependency modelings are less explored, and how to do theoretical analysis for them is an important research topic in the future. Copulas is an influential statistical tool for modeling dependence of multivariate data, and first brought into MLC for modeling label and feature dependencies \cite{DBLP:conf/NeurIPS/Liu19}. In particular, \cite{DBLP:conf/NeurIPS/Liu19} first constructs continuous distribution in the output space via employing the kernel trick, and then develops an unbiased and consistent estimator. Moreover, they also present the asymptotic analysis and mean squared error in the paper. However, the biggest problem for this paper is that it can not handle high dimension issues. The use of copula for modeling label and feature dependencies reveals new statistical insights in multi-label learning, and could orient more high dimension driven works in this direction.

\section{New Applications}\label{NewApplications}
During the past decade, multi-label classification has been successfully applied in various applications, such as protein function classification, music categorization and semantic scene classification. Recently, some new applications in computer vision (CV), natural language processing (NLP) and data mining (DM) are emerging, which are summarized in the Supplementary Materials. This section will briefly review some of them.

\subsection{Computer Vision}
\subsubsection{Video Annotation}
With the development of considerable videos on the Internet (e.g., Youtube, Flickr and Facebook), efficient and effective indexing and searching these video corpus becomes more and more important for the research and industry community. In many real-world video corpus, the videos are multi-labeled. For instance, most of the videos in the popular TRECVID data set \cite{DBLP:journals/pami/SnoekWGKSS06} are annotated by more than one label from a set of 39 different concepts.
Currently, semantic-level video annotation (i.e., the semantic video concept detection) has been an important research topic in the multimedia research community, which aims to tag videos with a set of concepts of interest, including scenes (e.g., garden, sky, tree), objects (e.g., animals, people, airplane, car), events (e.g., election, ceremony) and certain named entities (e.g., university, person, home).
\cite{orrelativeMultiLabelVideoAnnotation} attempts to capture the correlations between different labels to improve the annotation performance on video concepts. \cite{hua2008online} proposes a novel online multi-label learning method for large-scale video annotation.

\subsubsection{Facial Action Unit Recognition}
Thoughts and feelings are revealed in the face. 
The facial muscle movements tell a person's social behavior, psychopathology and internal states. Facial Action Unit (AU) Recognition plays an important role in describing comprehensive facial expressions, and has been successfully applied in mental state analysis.
Some works \cite{DBLP:conf/iccv/WangLWJ13} have provided the evidence that the occurrence of AUs are strongly correlated, and the sample distribution of AUs is unbalanced. Based on these properties, multi-label learning methods are well-matched to this learning scenario. For example, \cite{DBLP:conf/cvpr/ZhaoCTCZ15} introduces joint-patch and multi-label learning (JPML) to leverage group sparsity by selecting a sparse subset of facial patches while learning a multi-label classifier. \cite{DBLP:conf/cvpr/ZhaoCZ16} presents deep region and multi-label learning (DRML) for AU detection. Recently, \cite{DBLP:conf/NeurIPS/NiuHSC19} proposes a semi-supervised multi-label approach for AU recognition utilizing a large number of web face images without AU labels.

\subsubsection{Neonatal Brains}
Effective and consistent segmentation of brain white matter bundles at neonatal stage plays a vital role in detecting white matter abnormalities and understanding brain development for the prediction of psychiatric disorders. Because of the complexity of white matter anatomy and the spatial resolution of diffusion-weighted MR imaging, multiple fiber bundles can pass through one voxel. \cite{DBLP:journals/neuroimage/RatnarajahQ14} aims to assign one or multiple anatomical labels of white matter bundles to each voxel to reflect complex white matter anatomy of the neonatal brain. To achieve this goal, \cite{DBLP:journals/neuroimage/RatnarajahQ14} explores the supervised multi-label learning algorithm in Riemannian diffusion tensor spaces, which considers diffusion tensors lying on the Log-Euclidean Riemannian manifold of symmetric positive definite (SPD) matrices and their corresponding vector space as feature space. \cite{DBLP:journals/neuroimage/RatnarajahQ14} demonstrates that they are able to automatically learn the number of white matter bundles at a location and provide anatomical annotation of the neonatal white matter.
Recently, \cite{DBLP:journals/bspc/NoorizadehKDA19} and \cite{DBLP:journals/mta/NoorizadehKDBA20} present some weakly-supervised multi-label learning methods for neonatal brain extraction.
%
\subsection{Natural Language Processing}
\subsubsection{Mobile Applications}
Recently, the development of mobile applications has become one of the most important topics in communications \cite{DBLP:journals/cm/WunderJKWSCBGMFMCKDPEVW14}. Under this field, advanced high performance algorithms for mobile applications have attracted the attention of researchers. Recommendation systems are widely used to predict the ``rating'' or ``preference'' that a user would give to an item. A good recommendation system with high performance is able to attract users to the service for 5G applications. \cite{DBLP:journals/access/GuoJYYSH16} focuses on high performance multi-label classification methods and their applications for medical recommendations in the domain of 5G communication. \cite{DBLP:conf/icassp/Wang0XHDS20} develops a deep convolutional neural network for iris segmentation of noisy images acquired by mobile devices. A novel multi-label active learning method is proposed by \cite{DBLP:conf/ksem/MessaoudJJM19} for mobile reviews classification tasks. Mobile applications involve language understandings, we group it to NLP.

\subsubsection{Legal Text Mining}
MLC has been widely used in the legal domain, especially for legal text mining tasks. In 2008, Menc\'{\i}a and F\"{u}rnkranz \cite{DBLP:conf/lrec/MenciaF10} collects a data set EUR-Lex, which comprises of documents about European Union law, including treaties, legislation, case-law and legislative proposals. The documents are categorized into several orthogonal concepts according to the European Vocabulary (EUROVOC), to allow for multiple search facilities. Recently, there arises new interest. In \cite{DBLP:conf/acl/ChalkidisFMA19}, a new legal MLC data set, dubbed EURLEX-57K is released. This is a large-scale version of EUR-Lex data set (19.6k documents, 4k EUROVOC labels) that contains 57k EU legislative documents from the EUR-Lex portal, each of which is labeled by ~4.3k concepts from EUROVOC. In the Chinese AI and Law challenge \cite{DBLP:journals/corr/abs-1807-02478},
MLC is also applied to the legal judgment prediction (LJP) task, which aims to empower the machine to predict the judgment results of legal cases after reading fact descriptions. Since each criminal case can be relevant to multiple law articles, charges and prison terms, the LJP task can be regarded as a multi-label text classification problem. XMLC \cite{DBLP:conf/acl/ChalkidisFMA19} and DL MLC \cite{DBLP:journals/corr/abs-1807-02478} are proposed to address this task. Based on syntactic and grammatical features, legal text mining is categorized as NLP.

\subsection{Data Mining}
\subsubsection{Recommender Systems}
The recommender system can be naturally regarded as an MLC tasks since we usually recommend multiple items simultaneously to the users. For example, \cite{DBLP:conf/www/AgrawalGPV13} develops an MLC model to automatically recommend bid phrases to an advertiser from a given ad landing page; \cite{DBLP:conf/kdd/McAuleyPL15} approaches the item-to-item recommendation task on Amazon, which aims at predicting the subset of items (labels) that a user might buy along with a given item. A recent work \cite{DBLP:conf/kdd/ChangYZYD20} regarded the keyword recommendation as an XMLC task, that provides keyword suggestions for advertisers to create campaigns. The MLC model receives the product-query customer purchase records and then suggests queries that are relevant to any given product by utilizing product information, like title, description, brand, and so on. The applications of XMLC in recommendation have been widely studied in the literature \cite{DBLP:conf/www/AgrawalGPV13,DBLP:conf/kdd/McAuleyPL15,DBLP:conf/kdd/ChangYZYD20}.

\subsubsection{User Profiling}
In many applications, such as social media and e-commerce, it is essential to provide adaptive and personalized services to users. Therefore, user profiling, which infers user characteristics and personal interests from user-generated data, has been widely adopted by many online platforms. Some works regard this problem as a single-label learning task.
However, obviously, more user characteristics lead to better personalization and the correlations between different user profiles can help improve the quality of user profiling. Hence, some works try to infer multiple attributes simultaneously. For example, Farnadi \cite{DBLP:conf/wsdm/FarnadiTCM18} proposes a hybrid deep learning framework to infer multiple types of user-profiles from multiple modalities of user data.
Their experiments on 5K Facebook users also validates the superiority of the multi-label learning fashion to single-label learning. \cite{DBLP:conf/iccS/WenWZHG20} explores the user profiles on Weibo, a famous social network platform in China, by using graph information in social networks. Another example is fraud detection in e-commerce platforms \cite{DBLP:conf/ijcai/Wang2020}, since fraud users usually have different spam behaviors simultaneously. \cite{DBLP:conf/ijcai/Wang2020} presents a collaboration based multi-label propagation method to utilized the correlations among different fraud behaviors.

\section{Conclusion}
Multi-label classification has attracted significant attention from the community over the last decade. This paper provides a comprehensive review
of the emerging topics of multi-Label learning, which include extreme multi-label classification, multi-label learning with limited supervision, deep learning for multi-label learning, online multi-label learning, statistical multi-label learning and new applications. We provide an overview of the representative works referenced throughout. In addition, we emphasize the challenges of these emerging topics and some future research directions and the promising extensions that are worthy of further study.


%

\appendices
\section{Evaluation Metrics and Notations and New Applications}

\begin{table}[t]
\tiny
\caption{Important notations used in the main paper.}
\label{notations}
\begin{center}
\begin{tabular}{c|c}
\hline
Notations  & Explainations\\
\hline
$x_i,y_i$ & Input and output vectors \\
$X,Y$ & Input and output matrices \\
$\tilde{\mathcal{D}}$ & Transformed data set\\
$\hat{Y},\tilde{Y}$ & Label matrices of implicit and explicit missing labels\\
$Z=[z_1,\ldots,z_n]$ & Embedding matrix and vectors\\
$\dot{y}$ & Predicted score vector \\
$\breve{Y}, \breve{y}$ & Predicted logical label matrix and vector \\
$\Upsilon(y_i)$ & The indices of the positive labels of $y_i$\\
$\Lambda=[\lambda_{ij}]$ & Enriched real-value label representation \\
$S$ & The candidate label set in PML \\
$\Omega$ & The index set of neighbors\\
$\mathcal {N}_i$ & The index set of neighbors of the $i$-the instance\\
$D_l$, $D_o, D_u$& The index sets of labeled, incompletely-labeled and unlabeled data \\
$\bigtriangleup$ & The symmetric difference between two sets \\
$|\cdot|$ & The set cardinality \\
$\langle \cdot,\cdot \rangle$ & Inner product\\
$O(\cdot)$ & Computational complexity\\
$\cdot^T$ & Matrix Transpose \\
$\sigma(\cdot)$ & Sigmoid function\\
nnz$(\cdot)$ & The number of non-zero entries \\
$||\cdot ||_F,,||\cdot||_2,||\cdot||_1$ & Frobenius norm, $\ell_2$ and $\ell_1$ norm of a matrix (vector)\\
Tr$(\cdot)$ & Trace operator\\
$r(\cdot)$ & The regularizer function \\
$\mathcal{L}(\cdot)$ & Empirical risk function\\
$n$ &  The number of training data\\
$d,L$ & Feature dimensions and the number of labels\\
$\varpi$ & Dimension of embedding vectors\\
$\mathbb{R}$ & Set of real numbers \\
$s^*$ & The cardinality of the true support of the underlying model\\
$\alpha,\mu,\lambda,C$ & Trade-off hyperparameters \\
$I$ & Identity Matrix \\
$A,B$ & Side information matrices w.r.t input and output \\
$W,U,V,H$ & Projection or similarity Matrix \\
$F_e,F_x,F_d$ & Label encoding, feature encoding and decoding network of C2AE\\
$\mathcal{W}=[w_{ij}]_{n\times n}$ & Graph weight matrix\\
$L_x, L_y$ & The laplacian matrix of $\mathcal{W}^x$ and $\mathcal{W}^y$ \\
$\Phi(F_x, F_e), \Gamma(F_e, F_d)$ &  The losses of C2AE in the latent and output space \\
\hline
\end{tabular}
\end{center}
\vskip -0.3in
\end{table}

Assume $x_i \in \mathbb{R}^{d \times 1} $ is a real vector representing an input or instance (feature), $y_i=(y_{i,1},\cdots,y_{i,L}) \in \{0,1\}^{L \times 1}$ is the corresponding output or label vector $(i \in \{1,\ldots,n\})$. $n$, $d$ and $L$ denote the number of training data, feature dimensions and the number of labels, respectively. The input matrix is $X=[x_1,\ldots,x_n] \in \mathbb{R}^{d\times n} $ and the output matrix is $Y=[y_1,\ldots,y_n] \in \{0,1\}^{L \times n} $. MLC aims to learn a classifier which predicts the testing instance as accurate as possible with the set of proper labels. Let $\breve{Y}=[\breve{y}_1,\ldots,\breve{y}_n] \in \{0,1\}^{L \times n} $ be the predicted label. We first introduce some evaluation metrics for MLC.
\begin{table*}[t]
\caption{The new applications of multi-label learning.}
\label{TheapplicationsMLC}
\addtolength{\tabcolsep}{2.5pt}
\begin{center}
\tiny
\begin{tabular}{llll}
\multicolumn{1}{c}{\bf  Reference } & \multicolumn{1}{c}{\bf New Applications} & \multicolumn{1}{c}{\bf Approaches} & \multicolumn{1}{c}{\bf Evaluation Metrics}
\\ \hline \\
\cite{orrelativeMultiLabelVideoAnnotation}  & CV: automatic video annotation  & XMLC
\cite{DBLP:journals/ieeemm/NaphadeSTCHKHC06}, online MLC \cite{hua2008online} & Precision@$k$, Recall@$k$ and Hamming loss\\
\cite{DBLP:conf/aaai/ZhangSLL20} &CV: action recognition and localization in videos & multi-instance MLC \cite{DBLP:conf/aaai/ZhangSLL20}& Hamming loss\\
\cite{DBLP:conf/NeurIPS/NiuHSC19} &CV: facial action unit recognition & DL MLC \cite{DBLP:conf/cvpr/ZhaoCZ16}, semi-supervised MLC \cite{DBLP:conf/NeurIPS/NiuHSC19}&  Hamming loss and Ranking loss\\
\cite{DBLP:conf/NeurIPS/BucakJJ10}  &CV: visual object recognition & online MLC \cite{DBLP:conf/NeurIPS/BucakJJ10,DBLP:journals/corr/KimVV16}& Hamming loss \\
\cite{DBLP:journals/isci/HeGW12}  &CV: visual mobile robot navigation & multi-instance MLC \cite{DBLP:journals/isci/HeGW12} & Hamming loss \\
\cite{DBLP:conf/eccv/GradyF04}   &CV: biomedical image segmentation& semi-supervised MLC \cite{DBLP:conf/eccv/GradyF04}& Hamming loss\\
\cite{DBLP:journals/neuroimage/RatnarajahQ14}  &CV: neonatal brains &  semi-supervised MLC \cite{DBLP:journals/bspc/NoorizadehKDA19,DBLP:journals/mta/NoorizadehKDBA20}& Hamming loss \\
\cite{DBLP:journals/access/GuoJYYSH16}  &NLP: 5G mobile medical recommendations  & DL MLC \cite{DBLP:conf/icassp/Wang0XHDS20}, MLAL \cite{DBLP:conf/ksem/MessaoudJJM19}& Hamming loss and Ranking loss\\
\cite{DBLP:journals/is/SchulzMS16}  &NLP: social network analysis& DL MLC \cite{DBLP:conf/naacl/DongWHC19} &  Hamming loss and Ranking loss\\
\cite{DBLP:journals/corr/VenkatesanEDPW16}  &NLP: high-speed streaming data & online MLC \cite{DBLP:journals/corr/VenkatesanEDPW16} &  Hamming loss\\
\cite{DBLP:journals/nca/CiarelliOS14}  & NLP: web page categorization & DL MLC \cite{DBLP:journals/nca/CiarelliOS14}&  Hamming loss and Ranking loss\\
\cite{DBLP:conf/icassp/WanMZWK14}  & NLP: protein subcellular localization & XMLC \cite{DBLP:conf/icassp/WanMZWK14} & Precision@$k$ and Recall@$k$\\
\cite{DBLP:conf/icassp/WanMZWK14}  & NLP: legal text mining & XMLC \cite{DBLP:conf/acl/ChalkidisFMA19}, DL MLC \cite{DBLP:journals/corr/abs-1807-02478} & Precision@$k$, Recall@$k$ and Hamming loss\\
\cite{DBLP:conf/wsdm/FarnadiTCM18}  & DM: recommender system & XMLC \cite{DBLP:conf/www/AgrawalGPV13,DBLP:conf/kdd/McAuleyPL15,DBLP:conf/kdd/ChangYZYD20} & Precision@$k$, Recall@$k$, F-measure and Ranking loss\\
\cite{DBLP:conf/wsdm/FarnadiTCM18}  & DM: user profiling in social media & DL MLC\cite{DBLP:conf/wsdm/FarnadiTCM18}, semi-supervised MLC\cite{DBLP:conf/iccS/WeiZWLHH19}&  Hamming loss and Ranking loss\\
\cite{DBLP:conf/ijcai/Wang2020}   &  DM: e-commercial fraud user detection& semi-supervised MLC\cite{DBLP:conf/ijcai/Wang2020}&  Hamming loss and Ranking loss\\
\hline
\end{tabular}
\end{center}
\end{table*}

\textbf{Hamming loss.} Hamming loss is defined as follows:
\begin{equation*}\label{Hammingloss}
\begin{split}
1/n\sum_{i=1}^n |\Upsilon(y_i) \bigtriangleup \Upsilon(\breve{y}_i)|/L
\end{split}
\end{equation*}
where $\Upsilon(y_i)$ denotes the indices of the positive labels of $y_i$, $\bigtriangleup$ stands for the symmetric difference between two sets, $|\cdot|$ means the cardinality. The hamming loss evaluates the fraction
of misclassified instance-label pairs.

\textbf{Ranking loss.} Let $f$ be the real-valued function. Ranking loss is defined as follows:
\begin{equation*}\label{Rankingloss}
\begin{split}
1/n\!\sum_{i=1}^n \frac{|\{(a,b)\!: \!f(x_i,a) \!\leq\! f(x_i,b),(a,b)\in \!\Upsilon(y_i)\!\times \!\bar{y_i}\}|}{|\Upsilon(y_i)||\bar{y_i}|}
\end{split}
\end{equation*}
where $\bar{y_i}$ is the complementary set of $\Upsilon(y_i)$ in the label space. The ranking loss evaluates the fraction of reversely ordered label pairs.

\textbf{F-measure.}
\begin{equation*}\label{F-measure}
\begin{split}
\text{F-measure}=1/L\sum_{j=1}^L \frac{2\sum_{i=1}^n y_{i,j}\breve{y}_{i,j}}{\sum_{i=1}^n y_{i,j}+ \sum_{i=1}^n\breve{y}_{i,j}}
\end{split}
\end{equation*}
F-measure computes true positives, true negatives, false positives and false negatives over labels, and then calculates an overall F-1 score.

\textbf{Precision@$k$.}
\begin{equation*}\label{Precision@$k$}
\begin{split}
\text{Precision@}k=1/k \sum_{k\in \text{rank}_k(\dot{y})}y_k
\end{split}
\end{equation*}
where $\dot{y} \in \mathbb{R}^{L \times 1}$ is a predicted score vector, $y$ is a ground truth label vector and $rank_k(\dot{y})$ returns the $k$ largest indices of $\dot{y}$ ranked in descending order.

\textbf{Recall@$k$.}
\begin{equation*}\label{Recall@$k$}
\begin{split}
\text{Recall@}k=1/|\Upsilon(y)| \sum_{k\in \text{rank}_k(\dot{y})}y_k
\end{split}
\end{equation*}
Precision@$k$ and Recall@$k$ evaluate top-$k$ precision and recall over labels respectively, and both of them are the standard measures for XMLC. F-measure and Ranking loss are usually used in recommender system. Some CV and NLP applications, such as facial action unit recognition and web page categorization, usually use the Hamming loss and Ranking loss as the performance metric. The important notations and new applications in the main paper are summarized in Tables \ref{notations} and \ref{TheapplicationsMLC}, respectively.

\ifCLASSOPTIONcompsoc
%
%

\ifCLASSOPTIONcaptionsoff
  \newpage
\fi



%

%

\bibliographystyle{IEEEtran}
\bibliography{IEEEabrv,BibBib2}

\begin{IEEEbiography}[{\includegraphics[width=1in,height=1.25in,clip,keepaspectratio]{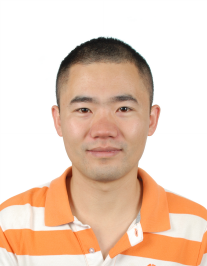}}]{Weiwei Liu}
received his PhD degree under the supervision of Prof. Ivor W. Tsang in computer science from University of Technology Sydney, Australia in 2017. He is currently a full professor with the School of Computer Science, Wuhan University, China. His current research interest is machine learning. His research results have been published at prestigious journals and leading conferences such as JMLR, IEEE TPAMI, IEEE TNNLS, IEEE TIP, IEEE TCYB, NeurIPS, ICML, ACL, AAAI, IJCAI and so on.
\end{IEEEbiography}

\begin{IEEEbiography}[{\includegraphics[width=1in,height=1.25in,clip,keepaspectratio]{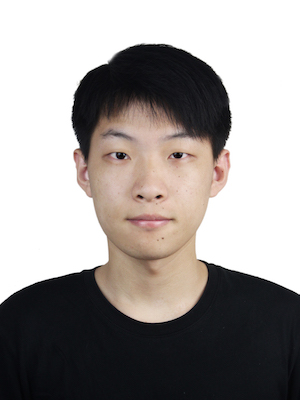}}]{Haobo Wang}
received his B.S. degree in Computer Science and Technology from Zhejiang University, China, in 2018. He is currently working toward the Ph.D. degree in the College of Computer Science and Technology, Zhejiang University. His research interests include machine learning and data mining, especially on multi-label learning and weakly-supervised learning.
\end{IEEEbiography}

\begin{IEEEbiography}[{\includegraphics[width=1in,height=1.25in,clip,keepaspectratio]{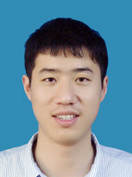}}]{Xiaobo Shen}
 received his BSc and PhD from School of Computer Science and Engineering, Nanjing University of Science and Technology in 2011 and 2017 respectively. He is currently a full professor with the School of Computer Science and Engineering, Nanjing University of Science and Technology, China. He has authored over 30 technical papers in prominent journals and conferences, such as IEEE TNNLS, IEEE TIP, IEEE TCYB, NeurIPS, ACM MM, AAAI, and IJCAI. His primary research interests are Multi-view Learning, Multi-label Learning, Network Embedding and Hashing.
\end{IEEEbiography}

\begin{IEEEbiography}[{\includegraphics[width=1in,height=1.25in,clip,keepaspectratio]{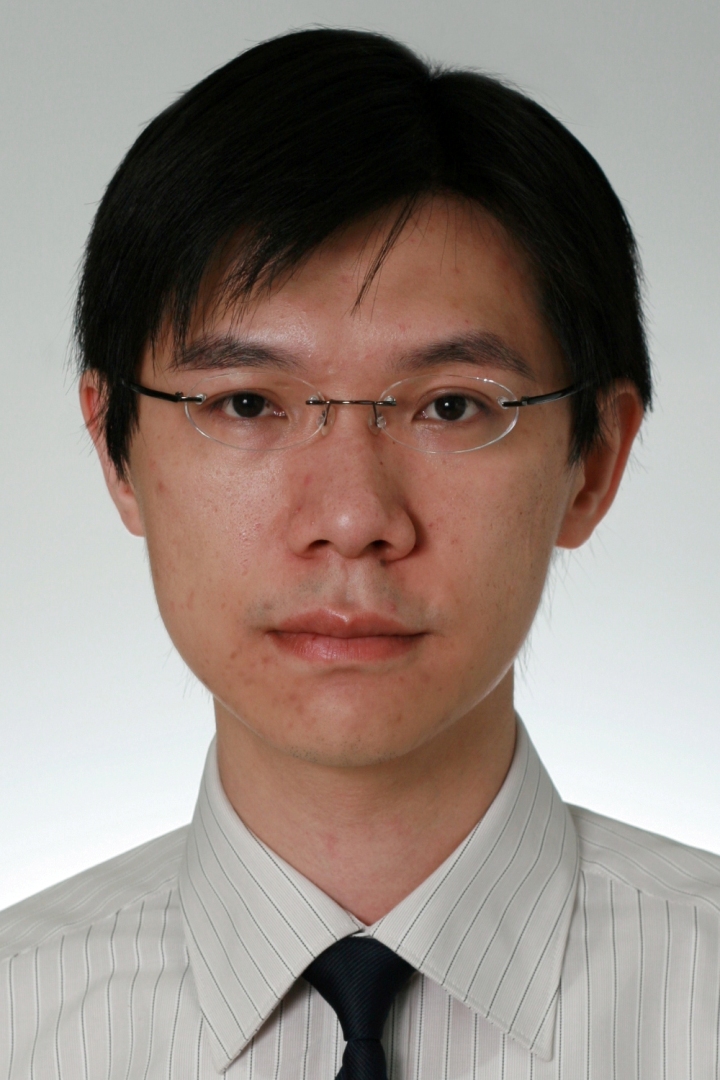}}]{Ivor W. Tsang}
 is an ARC Future Fellow and  Professor of Artificial Intelligence with the
University of Technology Sydney, Australia. He is also the Research Director of the
Australian Artificial Intelligence Institute.  His research interests include
transfer learning, generative models, and big data analytics for data with extremely high dimensions.  In 2013, Prof Tsang received his prestigious ARC Future Fellowship for his research regarding Machine Learning on Big Data. In 2019, his JMLR paper titled "Towards ultrahigh dimensional feature selection for big data" received the International Consortium of Chinese Mathematicians Best Paper Award. In 2020, Prof Tsang was recognized as the AI 2000 AAAI/IJCAI Most Influential Scholar in Australia for his outstanding contributions to the field of AAAI/IJCAI between 2009 and 2019. His research on transfer learning granted him the Best Student Paper Award at CVPR 2010 and the 2014 IEEE TMM Prize Paper Award. In addition, he received the IEEE TNN Outstanding 2004 Paper Award in 2007. He serves as a Senior Area Chair/Area Chair for NeurIPS, ICML, AISTATS, AAAI and IJCAI,  and the Editorial Board for JMLR, MLJ, and IEEE TPAMI.
\end{IEEEbiography}
\end{document}